# ANALYSIS OF BOOSTING ALGORITHMS USING THE SMOOTH MARGIN FUNCTION

BY CYNTHIA RUDIN[1,2,3], ROBERT E. SCHAPIRE[3,4]
AND INGRID DAUBECHIES[1,5]

*Columbia University, Princeton University and Princeton University*

We introduce a useful tool for analyzing boosting algorithms called the "smooth margin function," a differentiable approximation of the usual margin for boosting algorithms. We present two boosting algorithms based on this smooth margin, "coordinate ascent boosting" and "approximate coordinate ascent boosting," which are similar to Freund and Schapire's AdaBoost algorithm and Breiman's arc-gv algorithm. We give convergence rates to the maximum margin solution for both of our algorithms and for arc-gv. We then study AdaBoost's convergence properties using the smooth margin function. We precisely bound the margin attained by AdaBoost when the edges of the weak classifiers fall within a specified range. This shows that a previous bound proved by Rätsch and Warmuth is exactly tight. Furthermore, we use the smooth margin to capture explicit properties of AdaBoost in cases where cyclic behavior occurs.

**1. Introduction.** Boosting algorithms, which construct a "strong" classifier using only a training set and a "weak" learning algorithm, are currently among the most popular and most successful algorithms for statistical learning (see, e.g., Caruana and Niculescu-Mizil's recent empirical comparison of

Received October 2004; revised March 2007.
[1]Supported by NSF Grants DMS-98-10783 and DMS-02-19233 and University of Wisconsin Grant 640F161.
[2]Supported by an NSF Postdoctoral Research Fellowship and a grant from the Howard Hughes Medical Institute.
[3]Supported by NSF Grant IIS-03-25500.
[4]Supported by NSF Grant CCR-03-25463.
[5]Supported by AFOSR Grant F49620-01-0099.
This work was done while CR's affiliations were Program in Applied and Computational Mathematics and Department of Computer Science, Princeton University; and Center for Neural Science and Courant Institute of Mathematical Sciences, New York University.
*AMS 2000 subject classifications.* Primary 68W40, 68Q25; secondary 68Q32.
*Key words and phrases.* Boosting, AdaBoost, large margin classification, coordinate descent, arc-gv, convergence rates.

algorithms [3]). Freund and Schapire's AdaBoost algorithm [7] was the first practical boosting algorithm. AdaBoost maintains a discrete distribution (set of weights) over the training examples, and selects a weak classifier via the weak learning algorithm at each iteration. Training examples that were misclassified by the weak classifier at the current iteration then receive higher weights at the following iteration. The end result is a final combined classifier, given by a thresholded linear combination of the weak classifiers. See [13, 27] for an introduction to boosting.

Shortly after AdaBoost was introduced, it was observed that AdaBoost often does not seem to suffer from overfitting, in the sense that the test error does not go up even after a rather large number of iterations [1, 5, 14]. This lack of overfitting was later explained by Schapire et al. [28] in terms of the *margin theory*. The *margin* of a boosted classifier on a particular example is a number between $-1$ and $+1$ that can be interpreted as a measure of the classifier's confidence on this particular example. Further, the minimum margin over all examples in the training set is often referred to simply as the margin of the training set, or simply the margin when clear from context. Briefly, the margin theory states that AdaBoost tends to increase the margins of the training examples, and that this increase in the margins implies better generalization performance.

A complete analysis of AdaBoost's margin is nontrivial. Until recently, it was an open question whether or not AdaBoost always achieves the maximum possible margin. This question was settled (negatively) in [20, 22]; an example was presented in which AdaBoost's asymptotic margin was proved to be significantly below the maximum value. This example exhibited "cyclic" behavior, where AdaBoost's parameter values repeat periodically. So AdaBoost does not generally maximize the margin; furthermore, until the present work, the cyclic case was the only case for which AdaBoost's convergence was fully understood in the separable setting. When it cannot be proved that the parameters will eventually settle down into a cycle, AdaBoost's convergence properties are more difficult to analyze. Yet it seems essential to understand this convergence in order to study AdaBoost's generalization capabilities.

In this work, we introduce a new tool for analyzing AdaBoost and related algorithms. This tool is a differentiable approximation of the usual margin called the *smooth margin function*. We use it to provide the following main contributions.

- We identify an important new setting for which AdaBoost's convergence can be completely understood, called the case of *bounded edges*. A special case of our proof shows that the margin bound of Rätsch and Warmuth [17] is tight, closing what they allude to as a "gap in theory." This special case answers the question of how far below maximal AdaBoost's margin can be. Furthermore, this clarifies in sharp and precise terms the asymptotic relationship between the "edges" achieved by the weak learning algorithm and the asymptotic margin of AdaBoost.
- We derive two new algorithms similar to AdaBoost that are based directly on the smooth margin. Unlike AdaBoost, these algorithms provably converge to a maximum margin solution asymptotically; in addition, they possess a fast convergence rate to a maximum margin solution. Similar convergence rates based on the smooth margin are then presented for Breiman's arc-gv algorithm [2] answering what had been posed as an open problem by Meir and Rätsch [13].

1.1. *The case of bounded edges.* There is a rich literature connecting AdaBoost and margins. The margin theory of Schapire et al. [28] (later tightened by Koltchinskii and Panchenko [10]) showed that the larger the margins on the training examples, the better an upper bound on the generalization error, suggesting that, all else being equal, the generalization error can be reduced by systematically increasing the margins on the training set. Furthermore, Schapire et al. showed that AdaBoost has a tendency to increase the margins on the training examples. Thus, though not entirely complete, their theory and experiments strongly supported the notion that margins are highly relevant to the behavior and generalization performance of AdaBoost.

These bounds can be reformulated (in a slightly weaker form) in terms of the minimum margin; this was the focus of previous work by Breiman [2], Grove and Schuurmans [9] and Rätsch and Warmuth [17]. It is natural, given such an analysis, to pursue algorithms that will attempt to maximize this minimum margin. Such algorithms included Breiman's arc-gv algorithm [2] and Grove and Schuurmans' LP-AdaBoost [9] algorithm. However, in apparent contradiction of the margins theory, Breiman's experiments indicated that his algorithm achieved higher margins than AdaBoost, and yet performed worse on test data. Although this would seem to indicate serious trouble for the margins theory, recently, Reyzin and Schapire [18] revisited Breiman's experiments and were able to reconcile his results with the margins explanation, noting that the weak classifiers found by arc-gv are more complex than those found by AdaBoost. When this complexity is controlled, arc-gv continues to achieve larger minimum margins, but AdaBoost achieves much higher margins overall (and generally better test performance). Years earlier, Grove and Schuurmans [9] observed the same phenomenon; highly controlled experiments showed that AdaBoost achieved smaller minimum margins, overall larger margins, and often better test performance than LP-AdaBoost.

Taken together, these results indicate that there is a delicate and complex balance between the performance of the weak learning algorithm, the margins, the problem domain, the specific boosting algorithm being used, and the test error. It is the goal of the current work to improve our understanding of the intricate relationships between these various factors.

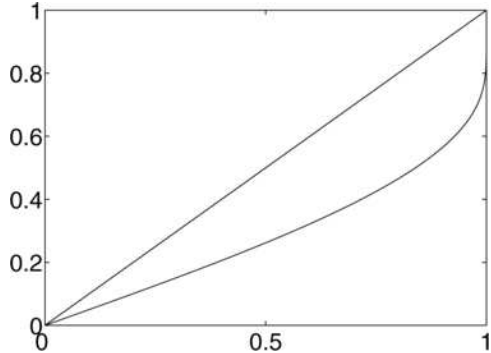

Fig. 1. *Plot of $\Upsilon(r)$ versus $r$ (lower curve), along with the function $f(r) = r$ (upper curve).*

In considering these complex relationships, a piece of the puzzle may be determined theoretically by understanding AdaBoost's convergence. AdaBoost has been shown to achieve large margins, but not maximal margins. To be precise, Schapire et al. [28] showed that AdaBoost achieves at least half of the maximum margin, that is, if the maximum margin is $\rho > 0$, AdaBoost will achieve a margin of at least $\rho/2$. This bound was tightened by Rätsch and Warmuth [17] who showed that AdaBoost asymptotically achieves a margin of at least $\Upsilon(\rho) > \rho/2$, where $\Upsilon : (0,1) \to (0,\infty)$ is the monotonically increasing function shown in Figure 1, namely,

$$\Upsilon(r) := \frac{-\ln(1-r^2)}{\ln((1+r)/(1-r))}. \tag{1.1}$$

However there is still a large gap between $\Upsilon(\rho)$ and the maximum margin $\rho$.

Our contribution is from the other direction; we have just described theoretical lower bounds for the margin, whereas we are now interested in upper bounds. Previously, we showed that it is possible for AdaBoost to achieve a margin that is significantly below the maximal value [22]. In this work, we show that Rätsch and Warmuth's bound is actually tight. In other words, we prove that it is possible for AdaBoost to achieve an asymptotic margin arbitrarily close to $\Upsilon(\rho)$. More generally, our theorem regarding the case of "bounded edges" says the following, where the "edge" measures the performance of the weak learning algorithm at each iteration:

- If AdaBoost's edge values are within a range $[\bar{\rho}, \bar{\rho}+\sigma]$ for some $\bar{\rho} \geq \rho$, then AdaBoost's margin asymptotically lies within the interval $[\Upsilon(\bar{\rho}), \Upsilon(\bar{\rho}+\sigma)]$.

Hence there is a fundamental connection between the performance of the weak learning algorithm and AdaBoost's asymptotic margin; if AdaBoost's edges fall within a given interval, we can find a corresponding interval for its asymptotic margin.

Now, since we have proven that we can more or less predetermine the value of AdaBoost's margin simply by specifying the edge values, we can perform a new experiment. Since the studies of Breiman [2] and Grove and Schuurmans [9] suggest that the margin theory cannot be easily tested using multiple algorithms, we now perform a controlled study with only one algorithm. The experiment in Section 7.2 consists of many trials with the same algorithm (AdaBoost) achieving different values of the margin on the same dataset. We find that as the (predetermined) margin increases, the probability of error on test data decreases dramatically. Our experiment supports the margin theory; in at least some cases, a larger margin does correlate with better generalization.

1.2. *Convergence properties of new and old algorithms.* Since AdaBoost may achieve a margin as low as $\Upsilon(\rho)$, and since it has the idiosyncratic (albeit fascinating and possibly helpful) tendency to sometimes get stuck in cyclic patterns [11, 22, 23], we are inspired to find algorithms that are similar to AdaBoost that have better convergence guarantees. We also study these cyclic patterns of AdaBoost as a special case for understanding its general convergence properties.

Our first main focus is to analyze two algorithms designed to maximize the smooth margin, called coordinate ascent boosting and approximate coordinate ascent boosting (presented in our previous work [23] without analysis). Coordinate ascent/descent algorithms are optimization algorithms where a step is made along only one coordinate at each iteration. The coordinate, which is also the choice of weak classifier, is determined by the weak learning algorithm. AdaBoost is also a coordinate descent algorithm [2, 6, 8, 12, 16], but its objective function need not be directly related to the margin or smooth margin; in fact, AdaBoost's objective converges to zero whenever the asymptotic margin is any positive value.

There are other algorithms designed to maximize the margin, though not based on coordinate ascent/descent of a fixed objective function. Here is a description of the known convergence properties of the relevant algorithms: AdaBoost does not converge to a maximum margin solution. Breiman's arc-gv algorithm [2, 13] has been proven to converge to the maximum margin asymptotically, but we are not aware of any proven convergence rate prior to this work. (Note that Meir and Rätsch [13] give a very simple asymptotic convergence proof for a variant of arc-gv; however, they note that no convergence rate can be derived from the proof.) Rätsch and Warmuth's AdaBoost$^*$ algorithm [17] has a fast convergence rate, namely, it yields a solution within $\bar{\nu}$ of the maximum margin in $2(\log_2 m)/\bar{\nu}^2$ steps, where $m$ is the number of training examples. However, the "greediness" parameter $\bar{\nu}$ must be manually entered (and perhaps adjusted) by the user; the algorithm is quite sensitive to $\bar{\nu}$. If it is estimated slightly too large or too small, the algorithm either takes a long time to converge, or it will not achieve the desired precision. (E.g., the experiments in [17] show that the algorithm performs well only for $\bar{\nu}$ in a carefully chosen range. In [25], $\bar{\nu}$ was estimated slightly too small, and the algorithm did not converge in a timely manner.) For any fixed value of $\bar{\nu}$, asymptotic convergence is not guaranteed and will generally not be achieved.

In contrast to previous algorithms, the ones we introduce have a proven fast convergence rate to the maximum margin, they have asymptotic convergence to the maximum margin, they do not require a choice of greediness parameter since the greediness is adaptively adjusted based on the progress of the algorithm, and they are based on coordinate ascent of a sensible objective, namely the smooth margin. The convergence rates for our algorithms and for arc-gv are custom-designed using recursive equalities for the smooth margin; we know of no standard techniques that would allow us to obtain such tight rates.

We also focus on the convergence properties of AdaBoost itself, using the smooth margin as a helpful analytical tool. The usefulness of the smooth margin follows largely from an important theorem, which shows that the value of the smooth margin increases if and only if AdaBoost takes a "large enough" step. Much previous work has focused on the statistical properties of AdaBoost indirectly through generalization bounds [10, 28], whereas our goal is to explore the way in which AdaBoost actually converges in order to produce a powerful classifier.

In Section 7.1, we use the smooth margin function to prove general properties of AdaBoost in cases where cyclic behavior occurs, extending previous work [22, 23]. "Cyclic behavior for AdaBoost" means that the weak learning algorithm repeatedly chooses the same sequence of weak classifiers, and the weight vectors repeat with a given period. When the number of training examples is small, it is likely that this behavior will be observed. Our first main result concerning cyclic AdaBoost is a proof that the value of the smooth margin must decrease an infinite number of times modulo one exception. Thus, a positive quality which holds for our new algorithms does not hold for AdaBoost: our new algorithms always increase the smooth margin at every iteration, whereas cyclic AdaBoost usually cannot. The single exception is the case where all edge values are identical. Our second result in this section concerns this exceptional case. We show that if all edges in a cycle are identical, then all support vectors (examples nearest the decision boundary) are misclassified by the same number of weak classifiers during the cycle. Thus, in this exceptional case, a strong equivalence exists between support vectors; they are misclassified the same proportion of the time by the weak learning algorithm.

Here is the outline for the full paper. In Section 2, we introduce our notation and explain the AdaBoost algorithm. In Section 3, we describe the smooth margin function that our algorithms are based on. In Section 4, we describe coordinate ascent boosting (Algorithm 1) and approximate coordinate ascent boosting (Algorithm 2), and in Section 5, the convergence of these algorithms is discussed, along with the convergence of arc-gv in Section 6. In Section 7, we show connections between AdaBoost and our smooth margin function. Specifically, in Section 7.1, we focus on cyclic AdaBoost, and in Section 7.2, we discuss the case of bounded edges, including the experiment described earlier. Sections 8, 9 and 10 contain proofs from Sections 3, 5, 6 and 7. Preliminary and less detailed statements of these results appear in [25, 26].

**2. Notation and introduction to AdaBoost.** Our notation is similar to that of Collins, Schapire and Singer [4]. The training set consists of examples with labels $\{(\mathbf{x}_i, y_i)\}_{i=1,\ldots,m}$, $m > 1$, where $(\mathbf{x}_i, y_i) \in \mathcal{X} \times \{-1, 1\}$. The space $\mathcal{X}$ never appears explicitly in our calculations. Let $\mathcal{H} = \{h_1, \ldots, h_n\}$ be the set of all possible weak classifiers that can be produced by the weak learning algorithm, where $h_j : \mathcal{X} \to \{-1, 1\}$. (The $h_j$'s are not assumed to be linearly independent; it is even possible that both $h$ and $-h$ belong to $\mathcal{H}$.) Since our classifiers are binary, and since we restrict our attention to their behavior on a finite training set, we can assume the number of weak classifiers $n$ is finite. We typically think of $n$ as being large, $m \ll n$, which makes a gradient descent calculation impractical; when $n$ is not large, the linear program can be solved directly using an algorithm such as LP-AdaBoost [9]. The classification rule that AdaBoost outputs is $f_{\text{Ada},\boldsymbol{\lambda}}$ where $\text{sign}(f_{\text{Ada},\boldsymbol{\lambda}})$ indicates the predicted class. The form of $f_{\text{Ada},\boldsymbol{\lambda}}$ is

$$f_{\text{Ada},\boldsymbol{\lambda}} := \frac{\sum_{j=1}^{n} \lambda_j h_j}{\|\boldsymbol{\lambda}\|_1},$$

where $\boldsymbol{\lambda} \in \mathbb{R}_+^n$ is the (unnormalized) coefficient vector. We define the 1-norm $\|\boldsymbol{\lambda}\|_1$ as usual: $\|\boldsymbol{\lambda}\|_1 := \sum_{j=1}^{n} \lambda_j$. At iteration $t$ of AdaBoost, the coefficient vector is $\boldsymbol{\lambda}_t$, and the sum is denoted $s_t := \|\boldsymbol{\lambda}_t\|_1$.

We define an $m \times n$ matrix $\mathbf{M}$ where $M_{ij} = y_i h_j(\mathbf{x}_i)$, that is, $M_{ij} = +1$ if training example $i$ is classified correctly by weak classifier $h_j$, and $-1$ otherwise. We assume that no column of $\mathbf{M}$ has all $+1$'s, that is, no weak classifier can classify all the training examples correctly. (Otherwise the learning problem is trivial.) This notation is useful mathematically for our analysis; however, it is not generally wise to explicitly construct large $\mathbf{M}$ in practice since the weak learning algorithm provides the necessary column for each iteration. $\mathbf{M}$ acts as the only "input" to AdaBoost in this notation, containing all the necessary information about the weak learning algorithm and training examples.

The margin theory developed via a set of generalization bounds that are based on the margin distribution of the training examples [10, 28], where the *margin of training example i* with respect to classifier $\boldsymbol{\lambda}$ is defined to be $y_i f_{\text{Ada},\boldsymbol{\lambda}}(\mathbf{x}_i)$, or equivalently, $(\mathbf{M}\boldsymbol{\lambda})_i/\|\boldsymbol{\lambda}\|_1$. These bounds can be reformulated (in a slightly weaker form) in terms of the minimum margin. We call the minimum margin over the training examples the *margin* of the training set, denoted $\mu(\boldsymbol{\lambda})$, that is,

$$\mu(\boldsymbol{\lambda}) := \min_i \frac{(\mathbf{M}\boldsymbol{\lambda})_i}{\|\boldsymbol{\lambda}\|_1}.$$

Any training example $i$ whose margin is equal to the minimum margin $\mu(\boldsymbol{\lambda})$ will be called a *support vector*. (There is a technical remark about our definition of AdaBoost. At iteration $t$, the (unnormalized) coefficient vector is denoted $\boldsymbol{\lambda}_t$; i.e., the coefficient of weak classifier $h_j$ determined by AdaBoost at iteration $t$ is $\lambda_{t,j}$. In the next iteration, all but one of the entries of $\boldsymbol{\lambda}_{t+1}$ are the same as in $\boldsymbol{\lambda}_t$; the only entry that is changed (for index $j = j_t$) is given a positive increment in our description of AdaBoost, i.e., $\lambda_{t+1,j_t} > \lambda_{t,j_t}$. Starting from $\boldsymbol{\lambda}_1 = \mathbf{0}$, this means that all the $\boldsymbol{\lambda}_t$ for $t > 1$ have nonnegative entries. We thus need to study the effect of AdaBoost only on the positive cone $\mathbb{R}^n_+ := \{\boldsymbol{\lambda} \in \mathbb{R}^n; \forall j \lambda_j \geq 0\}$. This same formalization was implicitly used in earlier works [17, 28]. Note that there are also formalizations; e.g., see [19], where entries of $\boldsymbol{\lambda}$ are permitted to decrease. The present formulation is also characterized by its focus on the coefficient vector $\boldsymbol{\lambda}$ as the "fundamental object," as opposed to the functional $\sum_j \lambda_j h_j$ defined by taking the $\lambda_j$ as weights for the $h_j$. This is expressed by our choice of the $\ell_1$ norm: $\|\boldsymbol{\lambda}\|_1 = \sum_j |\lambda_j|$ to "measure" $\boldsymbol{\lambda}$; if one focuses on the functional instead, then it is necessary to take into account that (because of the possible linear dependence of the $h_j$) several different choices of $\boldsymbol{\lambda}$ can give rise to the same functional. (E.g., if for some pair $\ell, \ell'$ we have $h_{\ell'} = -h_\ell$, then adding $\alpha$ to both $\lambda_\ell$ and $\lambda_{\ell'}$ does not change $\sum_j \lambda_j h_j$.) One must then use a norm that "quotients out" this ambiguity, as in (for instance) $\|\|\boldsymbol{\lambda}\|\| := \min\{\|\mathbf{a}\|_1; \sum_j a_j h_j = \lambda_j h_j\}$. By restricting ourselves to positive increments only, and using the $\ell_1$-norm of $\boldsymbol{\lambda}_t$, we avoid those nonunique issues. For our new algorithms, we prove $\lim_{t\to\infty}[\min_i(\mathbf{M}\boldsymbol{\lambda}_t)_i/\|\boldsymbol{\lambda}_t\|_1] = \rho$, and where $\rho$ is the maximum possible value of this quantity (defined later). Since $\|\boldsymbol{\lambda}_t\|_1 \geq \|\|\boldsymbol{\lambda}_t\|\|$, and $\rho$ is an upper bound for these fractions, it follows automatically that for our algorithms, $\lim_{t\to\infty}[\min_i(\mathbf{M}\boldsymbol{\lambda}_t)_i/\|\|\boldsymbol{\lambda}\|\|] = \rho$ as well; i.e., we prove convergence to a maximum margin solution even for the functional based norm. AdaBoost itself cannot be guaranteed to reach the maximum margin solution in the limit, regardless of whether $\|\boldsymbol{\lambda}_t\|_1$ or $\|\|\boldsymbol{\lambda}_t\|\|$ is used in the denominator.)

A boosting algorithm maintains a distribution, or set of weights, over the training examples that is updated at each iteration $t$. This distribution is denoted $\mathbf{d}_t \in \Delta_m$, and $\mathbf{d}_t^T$ is its transpose. Here, $\Delta_m$ denotes the simplex of $m$-dimensional vectors with nonnegative entries that sum to 1. At each iteration $t$, a weak classifier $h_{j_t}$ is selected by the weak learning algorithm. The probability of error at iteration $t$, denoted $d_-$, of the selected weak classifier $h_{j_t}$ on the training examples (weighted by the discrete distribution $\mathbf{d}_t$) is $d_- := \sum_{\{i\,:\,M_{ij_t}=-1\}} d_{t,i}$. Also, denote $d_+ := 1 - d_-$. Define $\mathcal{I}_+ := \{i : M_{ij_t} = +1\}$, the set of correctly classified examples at iteration $t$, and similarly define $\mathcal{I}_- := \{i : M_{ij_t} = -1\}$. Note that $d_+, d_-, \mathcal{I}_+$, and $\mathcal{I}_-$ depend on $t$; although we have simplified the notation, the iteration number will be clear from the context.

The *edge* of weak classifier $j_t$ at time $t$ is $r_t := (\mathbf{d}_t^T \mathbf{M})_{j_t} = d_+ - d_- = 1 - 2d_-$, with $(\cdot)_k$ indicating the $k$th vector component. Thus, a larger edge indicates a lower probability of error. Note that $d_+ = (1 + r_t)/2$ and $d_- = (1 - r_t)/2$. Also define

$$\gamma_t := \tanh^{-1} r_t = \frac{1}{2} \ln\left(\frac{1 + r_t}{1 - r_t}\right).$$

Due to the von Neumann Min–Max theorem for 2-player zero-sum games,

$$\min_{\mathbf{d} \in \Delta_m} \max_j (\mathbf{d}^T \mathbf{M})_j = \max_{\bar{\boldsymbol{\lambda}} \in \Delta_n} \min_i (\mathbf{M}\bar{\boldsymbol{\lambda}})_i.$$

That is, the minimum value of the maximum edge (left-hand side) corresponds to the maximum value of the margin. We denote this value by $\rho$.

We wish our learning algorithms to have robust convergence, so we will not generally require the weak learning algorithm to produce the weak classifier with the largest possible edge value at each iteration. Rather, we only require a weak classifier whose edge exceeds $\rho$, that is, $j_t \in \{j : (\mathbf{d}_t^T \mathbf{M})_j \geq \rho\}$. This notion of robustness has been previously used for the analysis of AdaBoost$^*$ and arc-gv. Here, AdaBoost in the *optimal* case means that the best weak classifier is chosen at every iteration: $j_t \in \arg\max_j (\mathbf{d}_t^T \mathbf{M})_j$, while AdaBoost in the *nonoptimal* case means that any good enough weak classifier is chosen: $j_t \in \{j : (\mathbf{d}_t^T \mathbf{M})_j \geq \rho\}$. The case of *bounded edges* is a subset of the nonoptimal case for some $\bar{\rho} \geq \rho$ and $\sigma \geq 0$, namely $j_t \in \{j : \bar{\rho} \leq (\mathbf{d}_t^T \mathbf{M})_j \leq \bar{\rho} + \sigma\}$.

We are interested in the *separable* case where $\rho > 0$ and the training error is zero; the margin specifically allows us to distinguish between classifiers that have zero training error. In the nonseparable case, AdaBoost's objective function $F$ is an upper bound on the training error, and convergence is well understood [4]. Not only does AdaBoost converge to the minimum of $F$, but it has recently been shown that it converges to the solution of the "bipartite

1. **Input:** Matrix $\mathbf{M}$, No. of iterations $t_{\max}$
2. **Initialize:** $\lambda_{1,j} = 0$ for $j = 1,\ldots,n$, also $d_{1,i} = 1/m$ for $i = 1,\ldots,m$, and $s_1 = 0$.
3. **Loop for** $t = 1,\ldots,t_{\max}$
   (a) $\left\{\begin{array}{l} j_t \in \operatorname{argmax}_j (\mathbf{d}_t^T \mathbf{M})_j \text{ optimal case} \\ j_t \in \{j : (\mathbf{d}_t^T \mathbf{M})_j \geq \rho\} \text{ nonoptimal case} \end{array}\right\}$
   (b) $r_t = (\mathbf{d}_t^T \mathbf{M})_{j_t}$
   (c) $g_t = \max[0, G(\boldsymbol{\lambda}_t)]$ where $G(\boldsymbol{\lambda}_t)$ is defined in (3.1), $G(\boldsymbol{\lambda}_t) = (-\ln \sum_{i=1}^m e^{-(\mathbf{M}\boldsymbol{\lambda}_t)_i})/s_t$.
   (d) $\left\{\begin{array}{ll} \alpha_t = \frac{1}{2} \ln\left(\frac{1+r_t}{1-r_t}\right) & \text{AdaBoost} \\ \alpha_t = \frac{1}{2} \ln\left(\frac{1+r_t}{1-r_t}\right) - \frac{1}{2} \ln\left(\frac{1+g_t}{1-g_t}\right) & \text{approx coord ascent boosting} \\ \text{If } g_t > 0, \alpha_t = \operatorname*{argmax}_\alpha G(\boldsymbol{\lambda}_t + \alpha \mathbf{e}_{j_t}), & \text{coord ascent boosting} \\ \text{else use AdaBoost.} & \end{array}\right\}$
   (e) $\boldsymbol{\lambda}_{t+1} = \boldsymbol{\lambda}_t + \alpha_t \mathbf{e}_{j_t}$, where $\mathbf{e}_{j_t}$ is 1 in position $j_t$ and 0 elsewhere.
   (f) $s_{t+1} = s_t + \alpha_t$
   (g) $d_{t+1,i} = d_{t,i} e^{-M_{ij_t} \alpha_t}/z_t$ where $z_t = \sum_{i=1}^m d_{t,i} e^{-M_{ij_t} \alpha_t}$
4. **Output:** $\boldsymbol{\lambda}_{t_{\max}}/s_{t_{\max}}$

FIG. 2. *Pseudocode for the AdaBoost algorithm, coordinate ascent boosting and approximate coordinate ascent boosting.*

ranking problem" at the same time; AdaBoost solves two problems for the price of one in the nonseparable case [21, 24]. However, in the separable case, where $F$ cannot distinguish between classifiers since it simply converges to zero, the margin theory suggests that we not only minimize $F$, but also distinguish between classifiers by choosing one that maximizes the margin. Since one does not know in advance whether the problem is separable, in this work we use AdaBoost until the problem becomes separable, and then perhaps switch to a mode designed explicitly to maximize the margin.

Figure 2 shows the pseudocode for AdaBoost, coordinate ascent boosting, and approximate coordinate ascent boosting. On each round of boosting, classifier $j_t$ with sufficiently large edge is selected (Step 3a), the weight of that classifier is updated (Step 3e), and the distribution $\mathbf{d}_t$ is updated and renormalized (Step 3g). Note that $\lambda_{t,j} = \sum_{\tilde{t}=1}^t \alpha_{\tilde{t}} \mathbf{1}_{j_{\tilde{t}}=j}$, where $\mathbf{1}_{j_{\tilde{t}}=j}$ is 1 if $j_{\tilde{t}} = j$ and 0 otherwise. The notation $\mathbf{e}_{j_t}$ means the vector that is 1 in position $j_t$ and 0 elsewhere.

2.1. *AdaBoost is coordinate descent.* AdaBoost is a coordinate descent algorithm for minimizing $F(\boldsymbol{\lambda}) := \sum_{i=1}^m e^{-(\mathbf{M}\boldsymbol{\lambda})_i}$. This has been shown many times [2, 6, 8, 12, 16], so we will only sketch the proof to introduce our notation. The direction AdaBoost chooses at iteration $t$ (corresponding to the choice of weak classifier $j_t$) in the optimal case is

$$j_t \in \arg\max_j \left[ -\frac{dF(\boldsymbol{\lambda}_t + \alpha \mathbf{e}_j)}{d\alpha} \bigg|_{\alpha=0} \right] = \arg\max_j \sum_{i=1}^{m} e^{-(\mathbf{M}\boldsymbol{\lambda}_t)_i} M_{ij}$$
$$= \arg\max_j (\mathbf{d}_t^T \mathbf{M})_j.$$

The step size AdaBoost chooses at iteration $t$ is $\alpha_t$, where $\alpha_t$ satisfies the following equation, that is, the equation for the line search along direction $j_t$:

$$0 = -\frac{dF(\boldsymbol{\lambda}_t + \alpha_t \mathbf{e}_{j_t})}{d\alpha_t} = \sum_{i=1}^{m} e^{-(\mathbf{M}(\boldsymbol{\lambda}_t + \alpha_t \mathbf{e}_{j_t}))_i} M_{ij_t},$$
$$0 = d_+ e^{-\alpha_t} - d_- e^{\alpha_t},$$
$$\alpha_t = \frac{1}{2} \ln\left(\frac{d_+}{d_-}\right) = \frac{1}{2} \ln\left(\frac{1+r_t}{1-r_t}\right) = \tanh^{-1} r_t = \gamma_t.$$

Note that for both the optimal and nonoptimal cases, $\alpha_t \geq \tanh^{-1} \rho > 0$, by monotonicity of $\tanh^{-1}$.

In the nonseparable case, the $\mathbf{d}_t$'s converge to a fixed vector [4]. In the separable case, the $\mathbf{d}_t$'s cannot converge to a fixed vector, and the minimum value of $F$ is 0, occurring as $\|\boldsymbol{\lambda}\|_1 \to \infty$. It is important to appreciate that this tells us nothing about the value of the margin achieved by AdaBoost or any other procedure designed to minimize $F$. In fact, an arbitrary algorithm that minimizes $F$ can achieve an arbitrarily bad (small) margin. [To see why, consider any $\bar{\boldsymbol{\lambda}} \in \Delta_n$ such that $(\mathbf{M}\bar{\boldsymbol{\lambda}})_i > 0$ for all $i$, assuming we are in the separable case so such a $\bar{\boldsymbol{\lambda}}$ exists. Then $\lim_{a \to \infty} a\bar{\boldsymbol{\lambda}}$ will produce a minimum value for $F$, but the original normalized $\bar{\boldsymbol{\lambda}}$ need not yield a maximum margin.] So it must be the *process* of coordinate descent that awards AdaBoost its ability to increase margins, not simply AdaBoost's ability to minimize $F$. The value of the function $F$ tells us very little about the value of the margin; even asymptotically, it only tells us whether the margin is positive or not.

A helpful property of AdaBoost is that we can do the line search at each step explicitly; that is, we have an analytical expression for the value of $\alpha_t$ for each $t$. Our second boosting algorithm, approximate coordinate ascent boosting, which incorporates an approximate line search, also has an update that can be solved explicitly.

**3. The smooth margin function $G(\boldsymbol{\lambda})$.** We wish to consider a function that, unlike $F$, actually tells us about the value of the margin. Our new function $G$ has the nice property that its maximum value corresponds to the maximum value of the margin. Here, $G$ is defined for $\boldsymbol{\lambda} \in \mathbb{R}_+^n$, $\|\boldsymbol{\lambda}\|_1 > 0$ by

$$G(\boldsymbol{\lambda}) := \frac{-\ln F(\boldsymbol{\lambda})}{\|\boldsymbol{\lambda}\|_1} = \frac{-\ln(\sum_{i=1}^m e^{-(\mathbf{M}\boldsymbol{\lambda})_i})}{\sum_j \lambda_j}. \tag{3.1}$$

One can think of $G$ as a smooth approximation of the margin, since it depends on the entire margin distribution when $\|\boldsymbol{\lambda}\|_1$ is small, and weights training examples with small margins much more highly than examples with larger margins, especially as $\|\boldsymbol{\lambda}\|_1$ grows. The function $G$ also bears a resemblance to the objective implicitly used for $\varepsilon$-boosting [19]. $G$ has many nice properties that are useful for understanding its geometry:

PROPOSITION 3.1 (Properties of the smooth margin [25]).

1. $G(\boldsymbol{\lambda})$ is a concave function (but not necessarily strictly concave) in each "shell" where $\|\boldsymbol{\lambda}\|_1$ is fixed.
2. The value of $G(\boldsymbol{\lambda})$ increases radially, that is, $G(a\boldsymbol{\lambda}) > G(\boldsymbol{\lambda})$ for $a > 1$.
3. As $\|\boldsymbol{\lambda}\|_1$ becomes large, $G(\boldsymbol{\lambda})$ tends to $\mu(\boldsymbol{\lambda})$. Specifically,

$$-\frac{\ln m}{\|\boldsymbol{\lambda}\|_1} + \mu(\boldsymbol{\lambda}) \leq G(\boldsymbol{\lambda}) < \mu(\boldsymbol{\lambda}).$$

PROOF. It follows from properties 2 and 3 that the maximum value of $G$ is the maximum value of the margin.

The proofs of properties 1 and 2 are in Section 8. Oddly enough, a lack of concavity does not affect our analysis, as our algorithms will iteratively maximize $G$, whether or not it is concave. For the proof of property 3,

$$me^{-\mu(\boldsymbol{\lambda})\|\boldsymbol{\lambda}\|_1} = \sum_{i=1}^m e^{-\min_\ell (\mathbf{M}\boldsymbol{\lambda})_\ell} \geq \sum_{i=1}^m e^{-(\mathbf{M}\boldsymbol{\lambda})_i}$$
$$> e^{-\min_i (\mathbf{M}\boldsymbol{\lambda})_i} = e^{-\mu(\boldsymbol{\lambda})\|\boldsymbol{\lambda}\|_1},$$

and taking logarithms, dividing by $\|\boldsymbol{\lambda}\|_1$ and negating yields the result. □

Since all values of the edge (even in the nonoptimal case) are required to be larger than the maximum margin $\rho$, we have for each iteration $t$, where recall $s_t := \|\boldsymbol{\lambda}_t\|_1$,

$$-\frac{\ln m}{s_t} + \mu(\boldsymbol{\lambda}_t) \leq G(\boldsymbol{\lambda}_t) < \mu(\boldsymbol{\lambda}_t) \leq \rho \leq r_t. \tag{3.2}$$

**4. Derivation of algorithms.** We now suggest two boosting algorithms that aim to maximize the margin explicitly (like arc-gv and AdaBoost$^*$), are based on coordinate ascent and adaptively adjust their step sizes (like AdaBoost). Before we derive the algorithms, we will write recursive equations for $F$ and $G$. This will provide a method for computing the values of $F$ and $G$ at iteration $t+1$ in terms of their values at iteration $t$. The recursive equation for $F$ is

$$F(\boldsymbol{\lambda}_t + \alpha \mathbf{e}_{j_t})$$
$$= \sum_{i=1}^m e^{-(\mathbf{M}(\boldsymbol{\lambda}_t + \alpha \mathbf{e}_{j_t}))_i} = \sum_{i \in \mathcal{I}_+} e^{-(\mathbf{M}\boldsymbol{\lambda}_t)_i} e^{-\alpha} + \sum_{i \in \mathcal{I}_-} e^{-(\mathbf{M}\boldsymbol{\lambda}_t)_i} e^{\alpha}$$
$$= [d_+ e^{-\alpha} + d_- e^{\alpha}] F(\boldsymbol{\lambda}_t) = \left[\frac{1+r_t}{2} e^{-\alpha} + \frac{1-r_t}{2} e^{\alpha}\right] F(\boldsymbol{\lambda}_t)$$
$$= [\cosh \alpha - r_t \sinh \alpha] F(\boldsymbol{\lambda}_t).$$

Here we remind the reader that $\cosh x = (e^x + e^{-x})/2$, $\sinh x = (e^x - e^{-x})/2$, and so $\cosh(\tanh^{-1} x) = (1 - x^2)^{-1/2}$. Recall the definition $\gamma_t := \tanh^{-1} r_t$. Continuing to reduce, we find the recursive equation for $F$,

$$F(\boldsymbol{\lambda}_t + \alpha \mathbf{e}_{j_t}) = \frac{\cosh \gamma_t \cosh \alpha - \sinh \gamma_t \sinh \alpha}{\cosh \gamma_t} F(\boldsymbol{\lambda}_t)$$

(4.1)

$$= \frac{\cosh(\gamma_t - \alpha)}{\cosh \gamma_t} F(\boldsymbol{\lambda}_t).$$

Here we have used the identity $\cosh(x-y) = \cosh x \cosh y - \sinh x \sinh y$. Now we find a recursive equation for $G$. By definition of $G$, we know $-\ln F(\boldsymbol{\lambda}_t) = s_t G(\boldsymbol{\lambda}_t)$. Taking the logarithm of (4.1) and negating,

$$(s_t + \alpha) G(\boldsymbol{\lambda}_t + \alpha \mathbf{e}_{j_t}) = -\ln F(\boldsymbol{\lambda}_t + \alpha \mathbf{e}_{j_t})$$
$$= -\ln F(\boldsymbol{\lambda}_t) - \ln\left(\frac{\cosh(\gamma_t - \alpha)}{\cosh \gamma_t}\right)$$

(4.2)

$$= s_t G(\boldsymbol{\lambda}_t) + \ln\left(\frac{\cosh \gamma_t}{\cosh(\gamma_t - \alpha)}\right)$$
$$= s_t G(\boldsymbol{\lambda}_t) + \int_{\gamma_t - \alpha}^{\gamma_t} \tanh u \, du.$$

Thus, we have a recursive equation for $G$. We will derive two algorithms; in the first, we assign to $\alpha_t$ the value $\alpha$ that maximizes $G(\boldsymbol{\lambda}_t + \alpha \mathbf{e}_{j_t})$, which requires solving an implicit equation. In the second algorithm, we pick an approximate value for the maximizer that can be computed in a straightforward way. In both cases, since it is not known in advance whether the problem is separable, the algorithm starts by running AdaBoost until $G(\boldsymbol{\lambda})$ becomes positive, which eventually must happen (in the separable case) by the following:

PROPOSITION 4.1. *In the separable case (where $\rho > 0$), AdaBoost achieves a positive value for $G(\boldsymbol{\lambda}_t)$ for some iteration $t$.*

PROOF. For the iteration defined by AdaBoost (i.e., $\alpha_t = \gamma_t = \tanh^{-1} r_t$), we have from (4.1)

$$F(\boldsymbol{\lambda}_{t+1}) = F(\boldsymbol{\lambda}_t + \gamma_t \mathbf{e}_{j_t}) = \frac{1}{\cosh \gamma_t} F(\boldsymbol{\lambda}_t) = (1 - r_t^2)^{1/2} F(\boldsymbol{\lambda}_t)$$

$$\leq (1 - \rho^2)^{1/2} F(\boldsymbol{\lambda}_t).$$

Hence, by this recursion, $F(\boldsymbol{\lambda}_{t+1}) \leq (1 - \rho^2)^{t/2} F(\boldsymbol{\lambda}_1)$. It follows that exceeding at most

$$\frac{2 \ln F(\boldsymbol{\lambda}_1)}{-\ln(1 - \rho^2)} + 1$$

iterations, $F(\boldsymbol{\lambda}_t) < 1$ so that $G(\boldsymbol{\lambda}_t) = (-\ln F(\boldsymbol{\lambda}_t))/s_t > 0$. □

For convenience in distinguishing the two algorithms defined below, we denote $\boldsymbol{\lambda}_1^{[1]}, \ldots, \boldsymbol{\lambda}_t^{[1]}$ to be a sequence of coefficient vectors generated by Algorithm 1, and $\boldsymbol{\lambda}_1^{[2]}, \ldots, \boldsymbol{\lambda}_t^{[2]}$ to be a sequence generated by Algorithm 2. Similarly, we distinguish the sequences $\alpha_t^{[1]}$ from $\alpha_t^{[2]}$, $g_t^{[1]} := G(\boldsymbol{\lambda}_t^{[1]})$, $g_t^{[2]} := G(\boldsymbol{\lambda}_t^{[2]})$, $s_t^{[1]} := \sum_j \lambda_{t,j}^{[1]}$ and $s_t^{[2]} := \sum_j \lambda_{t,j}^{[2]}$. Sometimes we compare the behavior of Algorithms 1 and 2 based on one iteration (from $t$ to $t+1$) as if they had started from the same coefficient vector at iteration $t$; we denote this vector by $\boldsymbol{\lambda}_t$. When an equation holds for both Algorithm 1 and Algorithm 2, we will often drop the superscripts. Although sequences such as $j_t$, $r_t$, $\gamma_t$, and $d_t$ are also different for Algorithms 1 and 2, we leave the notation without the superscript.

Note that it is important to compute $g_t$ in a numerically stable way. The pseudocode in Figure 2 might thus be replaced with

$$G(\boldsymbol{\lambda}_t) = \mu(\boldsymbol{\lambda}_t) - \frac{\ln \sum_{i=1}^m e^{-[(\mathbf{M}\boldsymbol{\lambda}_t)_i - \min_{i'}(\mathbf{M}\boldsymbol{\lambda})_{i'}]}}{s_t},$$

$$\text{where } \mu(\boldsymbol{\lambda}_t) = \frac{\min_i (\mathbf{M}\boldsymbol{\lambda}_t)_i}{s_t}.$$

4.1. *Algorithm* 1: *coordinate ascent boosting.* Let us consider coordinate ascent on $G$. In what follows, we will use only positive values of $G$, as we have justified via Proposition 4.1. The choice of direction $j_t$ at iteration $t$ (in the optimal case) obeys

$$j_t \in \arg\max_j \frac{dG(\boldsymbol{\lambda}_t^{[1]} + \alpha \mathbf{e}_j)}{d\alpha}\bigg|_{\alpha=0}$$

$$= \arg\max_j \left[\frac{\sum_{i=1}^m e^{-(\mathbf{M}\boldsymbol{\lambda}_t^{[1]})_i} M_{ij}}{F(\boldsymbol{\lambda}_t^{[1]})}\right] \frac{1}{s_t^{[1]}} + \frac{\ln(F(\boldsymbol{\lambda}_t^{[1]}))}{(s_t^{[1]})^2}.$$

Of the two terms on the right, the second term does not depend on $j$, and the first term is simply a constant times $(\mathbf{d}_t^T \mathbf{M})_j$. Thus the same direction will be chosen here as for AdaBoost. The "nonoptimal" setting we define for this algorithm will be the same as AdaBoost's, so the weak learning algorithm (Step 3a) of Algorithm 1 will be the same as AdaBoost's.

To determine the step size, ideally we would like to maximize $G(\boldsymbol{\lambda}_t^{[1]} + \alpha \mathbf{e}_{j_t})$ with respect to $\alpha$, that is, we would like to define the step size $\alpha_t^{[1]}$ to obey $dG(\boldsymbol{\lambda}_t^{[1]} + \alpha \mathbf{e}_{j_t})/d\alpha = 0$ for $\alpha = \alpha_t^{[1]}$. Differentiating (4.2) gives

$$(s_t^{[1]} + \alpha)\frac{dG(\boldsymbol{\lambda}_t^{[1]} + \alpha \mathbf{e}_{j_t})}{d\alpha} + G(\boldsymbol{\lambda}_t^{[1]} + \alpha \mathbf{e}_{j_t}) = \tanh(\gamma_t - \alpha).$$

Thus, our ideal step size $\alpha_t^{[1]}$ satisfies

(4.3) $$G(\boldsymbol{\lambda}_{t+1}^{[1]}) = G(\boldsymbol{\lambda}_t^{[1]} + \alpha_t^{[1]} \mathbf{e}_{j_t}) = \tanh(\gamma_t - \alpha_t^{[1]}).$$

There is not a nice analytical solution for $\alpha_t^{[1]}$ (as there is for AdaBoost), but minimization of $G(\boldsymbol{\lambda}_t^{[1]} + \alpha \mathbf{e}_{j_t})$ is one-dimensional so it can be performed reasonably quickly. Hence we have defined the first of our new boosting algorithms, coordinate ascent on $G$, implementing a line search at each iteration. Furthermore:

PROPOSITION 4.2. *The solution for $\alpha_t^{[1]}$ is unique, for some $\alpha_t^{[1]} > 0$.*

PROOF. First, we rewrite the line search equation (4.3) using (4.2),

$$s_t^{[1]} G(\boldsymbol{\lambda}_t^{[1]}) + \ln\left(\frac{\cosh \gamma_t}{\cosh(\gamma_t - \alpha_t^{[1]})}\right) = (s_t^{[1]} + \alpha_t^{[1]})\tanh(\gamma_t - \alpha_t^{[1]}).$$

Consider the function $f_t$,

$$f_t(\alpha) := s_t^{[1]} G(\boldsymbol{\lambda}_t^{[1]}) + \ln\left(\frac{\cosh \gamma_t}{\cosh(\gamma_t - \alpha)}\right) - (s_t^{[1]} + \alpha)\tanh(\gamma_t - \alpha).$$

Now, $df_t(\alpha)/d\alpha = (\alpha + s_t^{[1]})\operatorname{sech}^2(\gamma_t - \alpha) > 0$ for $\alpha > 0$. Thus $f_t$ is strictly increasing, so there is at most one root. We also have $f_t(0) = s_t^{[1]}(G(\boldsymbol{\lambda}_t^{[1]}) -$

$r_t) < 0$ and $f_t(\gamma_t) = s_t^{[1]} G(\boldsymbol{\lambda}_t^{[1]}) - \frac{1}{2}\ln(1 - r_t^2) > 0$. Thus, by the intermediate value theorem, there is at least one root. Hence, there is exactly one solution for $\alpha_t^{[1]}$ where $\alpha_t^{[1]} > 0$.  □

Let us rearrange our equations slightly in order to study the update. Using the notation $g_{t+1}^{[1]} := G(\boldsymbol{\lambda}_{t+1}^{[1]})$ in (4.3), we find that $\alpha_t^{[1]}$ satisfies the following (implicitly):

$$\begin{aligned}(4.4)\quad \alpha_t^{[1]} &= \gamma_t - \tanh^{-1}(g_{t+1}^{[1]}) = \tanh^{-1} r_t - \tanh^{-1}(g_{t+1}^{[1]}) \\ &= \frac{1}{2}\ln\left[\frac{1+r_t}{1-r_t}\frac{1-g_{t+1}^{[1]}}{1+g_{t+1}^{[1]}}\right].\end{aligned}$$

Since $G(\boldsymbol{\lambda}_{t+1}^{[1]}) \geq G(\boldsymbol{\lambda}_t^{[1]})$, we again have $G(\boldsymbol{\lambda}_{t+1}^{[1]}) > 0$, and thus $\alpha_t^{[1]} \leq \tanh r_t = \gamma_t$. Hence, the step size for this new algorithm is always positive, and it is upper-bounded by AdaBoost's step size.

4.2. *Algorithm* 2: *approximate coordinate ascent boosting.* The second of our two new boosting algorithms avoids the line search of Algorithm 1, and is even slightly more aggressive. It seems to perform very similarly to Algorithm 1 in our experiments. To define this algorithm, we consider the following approximate solution to the maximization problem, by using an approximate solution to (4.3) at each iteration in which $\boldsymbol{\lambda}_{t+1}$ is replaced by $\boldsymbol{\lambda}_t$ for tractability:

$$(4.5)\qquad G(\boldsymbol{\lambda}_t^{[2]}) = \tanh(\gamma_t - \alpha_t^{[2]}),$$

or more explicitly,

$$\begin{aligned}(4.6)\quad \alpha_t^{[2]} &= \gamma_t - \tanh^{-1}(g_t^{[2]}) = \tanh^{-1} r_t - \tanh^{-1}(g_t^{[2]}) \\ &= \frac{1}{2}\ln\left[\frac{1+r_t}{1-r_t}\frac{1-g_t^{[2]}}{1+g_t^{[2]}}\right].\end{aligned}$$

The update $\alpha_t^{[2]}$ is also strictly positive, since $g_t^{[2]} < \rho \leq r_t$, by (3.2). Note that this choice for $\alpha_t^{[2]}$ given by (4.5) implies, by (4.2), using the monotonicity of tanh to take the lower endpoint on the integral,

$$(s_t^{[2]} + \alpha_t^{[2]})G(\boldsymbol{\lambda}_{t+1}^{[2]}) > s_t^{[2]} G(\boldsymbol{\lambda}_t^{[2]}) + \alpha_t^{[2]} \tanh(\gamma_t - \alpha_t^{[2]})$$
$$= (s_t^{[2]} + \alpha_t^{[2]})G(\boldsymbol{\lambda}_t^{[2]}),$$

so that $G(\boldsymbol{\lambda}_{t+1}^{[2]}) > G(\boldsymbol{\lambda}_t^{[2]})$. That is, Algorithm 2 still increases $G$ at every iteration. In particular, $G(\boldsymbol{\lambda}_{t+1}^{[2]})$ is again strictly positive.

Algorithm 2 is slightly more aggressive than Algorithm 1, in the sense that it picks a larger relative step size $\alpha_t$, albeit not as large as the step size defined by AdaBoost itself. We can see this by comparing equations (4.4) and (4.6). If Algorithms 1 and 2 were started at the same position $\boldsymbol{\lambda}_t$, with $g_t := G(\boldsymbol{\lambda}_t)$, then Algorithm 2 would always take a slightly larger step than Algorithm 1; since $g_{t+1}^{[1]} > g_t$, we have $\alpha_t^{[1]} < \alpha_t^{[2]}$.

**5. Convergence of smooth margin algorithms.** We will show convergence of Algorithms 1 and 2 to a maximum margin solution. Although there are many papers describing the convergence of specific classes of coordinate descent/ascent algorithms, this problem did not fit into any of the existing categories. For example, we were unable to fit our algorithms into any of the categories described by Zhang and Yu [29], but we did use some of their key ideas as inspiration for our proofs for this section, which can all be found in Section 9.

One of the main results of this analysis is that both algorithms make significant progress at each iteration. In the next lemma, we are only considering one increment, so we fix $\boldsymbol{\lambda}_t$ at iteration $t$ and let $g_t := G(\boldsymbol{\lambda}_t)$, $s_t := \sum_j \lambda_{t,j}$. Then, denote the next values of $G$ for Algorithms 1 and 2, respectively, as $g_{t+1}^{[1]} := G(\boldsymbol{\lambda}_t + \alpha_t^{[1]} \mathbf{e}_{j_t})$ and $g_{t+1}^{[2]} := G(\boldsymbol{\lambda}_t + \alpha_t^{[2]} \mathbf{e}_{j_t})$. Similarly, $s_{t+1}^{[1]} := s_t + \alpha_t^{[1]}$ and $s_{t+1}^{[2]} := s_t + \alpha_t^{[2]}$.

LEMMA 5.1 (Progress at every iteration).

$$g_{t+1}^{[1]} - g_t \geq \frac{\alpha_t^{[1]}(r_t - g_t)}{2s_{t+1}^{[1]}} \quad and \quad g_{t+1}^{[2]} - g_t \geq \frac{\alpha_t^{[2]}(r_t - g_t)}{2s_{t+1}^{[2]}}.$$

Another important ingredient for our convergence proofs is that the step size does not increase too quickly; this is the main content of the next lemma.

LEMMA 5.2 (Step size does not increase too quickly).

$$\lim_{t \to \infty} \frac{\alpha_t^{[1]}}{s_{t+1}^{[1]}} = 0 \quad and \quad \lim_{t \to \infty} \frac{\alpha_t^{[2]}}{s_{t+1}^{[2]}} = 0.$$

Lemmas 5.1 and 5.2 allow us to show convergence of Algorithms 1 and 2 to a maximum margin solution. Recall that for convergence, it is sufficient to show that $\lim_{t \to \infty} g_t = \rho$ since $g_t < \mu(\boldsymbol{\lambda}_t) \leq \rho$.

THEOREM 5.1 (Asymptotic convergence). *Algorithms 1 and 2 converge to a maximum margin solution, that is, $\lim_{t \to \infty} g_t^{[1]} = \rho$ and $\lim_{t \to \infty} g_t^{[2]} = \rho$. And thus, $\lim_{t \to \infty} \mu(\boldsymbol{\lambda}_t^{[1]}) = \rho$ and $\lim_{t \to \infty} \mu(\boldsymbol{\lambda}_t^{[2]}) = \rho$.*

Theorem 5.1 guarantees asymptotic convergence, without providing any information about a rate of convergence. In what follows, we shall state two different results about the convergence rate. The first theorem gives an explicit a priori upper bound on the number of iterations needed to guarantee that $g_t^{[1]}$ or $g_t^{[2]}$ is within $\varepsilon > 0$ of the maximum margin $\rho$. As is often the case for uniformly valid upper bounds, the convergence rate provided by this theorem is not optimal, in the sense that faster decay of $\rho - g_t$ can be proved for large $t$ if one does not insist on explicit constants. The second convergence rate theorem provides such a result, stating that $\rho - g_t = \mathcal{O}(t^{-1/(3+\delta)})$, or equivalently $\rho - g_t \leq \varepsilon$ after $\mathcal{O}(\varepsilon^{-(3+\delta)})$ iterations, where $\delta > 0$ can be arbitrarily small.

Both convergence rate theorems rely on estimates limiting the growth rate of $\alpha_t$. Lemma 5.2 is one such estimate; because it is only an asymptotic estimate, our first convergence rate theorem requires the following uniformly valid lemma.

LEMMA 5.3 (Step size bound).
$$\alpha_t^{[1]} \leq c_1 + c_2 s_t^{[1]} \quad and \quad \alpha_t^{[2]} \leq c_1 + c_2 s_t^{[2]},$$
*where*
$$c_1 = \frac{\ln 2}{1-\rho} \quad and \quad c_2 = \frac{\rho}{1-\rho}.$$

We are now ready for a first convergence rate theorem. We leave off superscripts when the statement is true for both algorithms.

THEOREM 5.2 (Convergence rate). *Let $\tilde{1}$ be the iteration at which $G$ becomes positive. Then both the margin $\mu(\boldsymbol{\lambda}_t)$ and the value of $G(\boldsymbol{\lambda}_t)$ will be within $\varepsilon$ of the maximum margin $\rho$ within at most*
$$\tilde{1} + (s_{\tilde{1}} + \ln 2)\varepsilon^{-(3-\rho)/(1-\rho)}$$
*iterations, for both Algorithms 1 and 2.*

In practice $\rho$ is unknown; this means one cannot use Theorem 5.2 directly in order to get an explicit numerical upper bound on the number of iterations required to achieve the given accuracy $\varepsilon$. However, if $R$ is an explicit upper bound on $\rho$, then the same argument can be used to prove that $g_t$ will exceed $\rho - \varepsilon$ within at most
$$\tilde{1} + (s_{\tilde{1}} + \ln 2)\varepsilon^{-(3-R)/(1-R)}$$
iterations. If $R$ is close to $\rho$, this bound becomes tighter. As we iterate, we can obtain increasingly better upper bounds $R_t$ on $\rho$ as follows: since we have assumed that the weak learning algorithm produces an edge of at least $\rho$, that is, $r_\ell \geq \rho$ for all $\ell$, it follows that $R_t := \min_{\ell \leq t} r_\ell$ is an upper bound for $\rho$. $R_t$ is known explicitly at iteration $t$ since the numerical values for all the $r_\ell$ where $\ell \leq t$ are known. We thus obtain, as a corollary to the proof of Theorem 5.2, the following result, valid for both algorithms.

COROLLARY 5.1. *Let $\tilde{1}$ be the iteration at which $G$ becomes positive. At any later iteration $t$, if the algorithms are continued for at most*

$$\Delta t := \tilde{1} + (s_{\tilde{1}} + \ln 2)\varepsilon^{-(3-R_t)/(1-R_t)} - t$$

*additional iterations, where $R_t = \min_{\ell \leq t} r_\ell$, then $g_{t+\Delta t} \in [\rho - \varepsilon, \rho]$.*

That is, the value of $G$ will be within $\varepsilon$ of the maximum margin $\rho$ in at most $\Delta t$ additional iterations. Note that if $\Delta t$ is negative, then we have already achieved $g_t \in [\rho - \varepsilon, \rho]$.

An important remark is that the technique of proof of Theorem 5.2 is much more widely applicable. In fact, we later use this framework to prove a convergence rate for arc-gv. The proof used only two main ingredients, Lemmas 5.1 and 5.3. Note that AdaBoost itself obeys Lemma 5.3; in fact, a bound of the same form can be seen solely from Lemma 5.3 and one additional fact, namely, starting from $\boldsymbol{\lambda}_t$, the step size $\alpha_t$ for AdaBoost only exceeds $\alpha_t^{[1]}$ and $\alpha_t^{[2]}$ by at most a constant, specifically $\frac{1}{2}\ln(\frac{1+\rho}{1-\rho})$. It is the condition of Lemma 5.1 that AdaBoost does not obey; AdaBoost does not make progress with respect to $G$ at each iteration as we discuss in Section 7.

The convergence rate provided by Theorem 5.2 is not tight; in fact, Algorithms 1 and 2 often perform at a much faster rate of convergence in practice. The fact that the step-size bound in Lemma 5.3 holds for all $t$ allowed us to find an upper bound on the number of iterations; however, we can find faster convergence rates in the asymptotic regime by using Lemma 5.2 instead. The following lemma again holds for both Algorithm 1 and Algorithm 2, and we drop the superscripts.

LEMMA 5.4. *For any $0 < \nu < 1/2$, there exists a constant $C_\nu$ such that for all $t \geq \tilde{1}$,*

$$\rho - g_t \leq C_\nu s_t^{-\nu}.$$

Let us turn this into a convergence rate estimate. Note that the big-oh notation in this theorem hides constants that depend on the matrix $\mathbf{M}$.

THEOREM 5.3 (Faster convergence rate). *For both Algorithms 1 and 2, and for any $\delta > 0$, a margin within $\varepsilon$ of optimal is obtained after at most $\mathcal{O}(\varepsilon^{-(3+\delta)})$ iterations from the iteration $\tilde{1}$ where $G$ becomes positive.*

Although Theorem 5.3 gives a better convergence rate than Theorem 5.2 [since $3 < (3-\rho)/(1-\rho)$], there is a constant factor that is not explicitly given. Hence, this estimate cannot be translated into an a priori upper bound on the number of iterations after which $\rho - g_t < \varepsilon$ is guaranteed, unlike Theorem 5.2 or Corollary 5.1.

From our experiments with Algorithms 1 and 2, we have noticed that they converge much faster than predicted (see [25]). This is especially true when the edges are large. Nevertheless, the asymptotic convergence rate of Theorem 5.3 is sharp in the most extreme nonoptimal case where the weak learning algorithm always achieves an edge of $\rho$, as shown in the following theorem. This theorem is proved for Algorithm 2 only, as it conveys our point and eases notation.

THEOREM 5.4 (Convergence rate is sharp). *Suppose $r_t = \rho$ for all $t$. Then, there exists no $C > 0$, $\delta > 0$, $t_0 > 0$ so that $\rho - g_t^{[2]} \leq Ct^{-(1/3)-\delta}$ for all $t \geq t_0$. Equivalently, for all $\delta > 0$, $\limsup_{t \to \infty} t^{1+\delta}(\rho - g_t^{[2]})^3 = \infty$, showing that Algorithm 2 requires at least $\Omega(\varepsilon^{-3})$ iterations to achieve a value of $g_t^{[2]}$ within $\varepsilon$ of optimal. That is, the convergence rate of Theorem 5.3 is sharp.*

**6. Convergence of arc-gv.** We have finished describing the smooth margin algorithms. We will now alter our course; we will use the smooth margin function to study well-known algorithms, first arc-gv and then AdaBoost. arc-gv is defined as in Figure 2 except that the update in Step 3d is replaced by $\alpha_t^{\mathrm{arc}}$,

$$\alpha_t^{\mathrm{arc}} = \frac{1}{2}\ln\left(\frac{1+r_t}{1-r_t}\right) - \frac{1}{2}\ln\left(\frac{1+\mu_t}{1-\mu_t}\right), \qquad \text{where } \mu_t := \mu(\boldsymbol{\lambda}_t).$$

(Note that we are using Breiman's original formulation of arc-gv, not Meir and Rätsch's variation.) Note that $\alpha_t^{\mathrm{arc}}$ is nonnegative since $\mu_t \leq \rho \leq r_t$. We directly present a convergence rate for arc-gv; most of the important computations for this bound have already been established in the proof of Theorem 5.2. As before, we start from when the smooth margin is positive. For arc-gv, the smooth margin increases at each iteration (and the margin does not necessarily increase). The result we state is weaker than the bound for Algorithms 1 and 2, since it is in terms of the maximum margin achieved up to time $t$ rather than in terms of the smooth margin at time $t$. However, we note that the smooth margin does increase monotonically, and the true margin is never far from the smooth margin as we have shown in Proposition 3.1. Here is our guaranteed convergence rate:

THEOREM 6.1 (Convergence rate for arc-gv). *Let $\tilde{1}$ be the iteration at which $G$ becomes positive. Then $\max_{\{\ell=\tilde{1},\ldots,t\}} \mu(\boldsymbol{\lambda}_\ell)$ will be within $\varepsilon$ of the* *maximum margin $\rho$ within at most*

$$\tilde{1} + (s_{\tilde{1}} + \ln 2)\varepsilon^{-(3-\rho)/(1-\rho)}$$

*iterations, for arc-gv.*

The proof is given in Section 9.

**7. A new way to measure AdaBoost's progress.** In many ways, AdaBoost is still a mysterious algorithm. Although it often seems to converge to a maximum margin solution (at least in the optimal case), it was shown via some optimal case examples that it does not always do so [20, 22]. In fact, the difference between the margin produced by AdaBoost and the maximum margin can be quite large; we shall see below that this happens when the edges are forced to be somewhat small. These and other results [2, 9, 22] suggest that the margin theory only provides a significant piece of the puzzle of AdaBoost's strong generalization properties; it is not the whole story. In order to understand AdaBoost's strong generalization abilities, it is essential to understand *how* AdaBoost actually constructs its solutions. In this section, we make use of new tools to help us understand how AdaBoost makes progress. Namely, we measure the progress of AdaBoost according to a quantity other than the margin, namely, the smooth margin function $G$. We focus on two cases: the case where AdaBoost cycles, and the case of bounded edges, where AdaBoost's edges are required to be bounded strictly below 1. These are the only cases for which AdaBoost's convergence is understood for separable data.

First, we show that whenever AdaBoost takes a large step, it makes progress according to $G$. This result will form the basis of all other results in this section. We will use the superscript $[A]$ for AdaBoost. Our analysis makes use of a monotonically increasing function $\Upsilon : (0,1) \to (0, \infty)$, which is defined as

$$\Upsilon(r) := \frac{-\ln(1-r^2)}{\ln(1+r)/(1-r)}.$$

One can show that $\Upsilon$ is monotonically increasing by considering its derivative. A plot of $\Upsilon$ is shown in Figure 1.

THEOREM 7.1 (AdaBoost makes progress if and only if it takes a large step).

$$G(\boldsymbol{\lambda}_{t+1}^{[A]}) \geq G(\boldsymbol{\lambda}_t^{[A]}) \quad \Longleftrightarrow \quad \Upsilon(r_t) \geq G(\boldsymbol{\lambda}_t^{[A]}).$$

In other words, $G(\boldsymbol{\lambda}_{t+1}^{[A]}) \geq G(\boldsymbol{\lambda}_t^{[A]})$ if and only if the edge $r_t$ is sufficiently large.

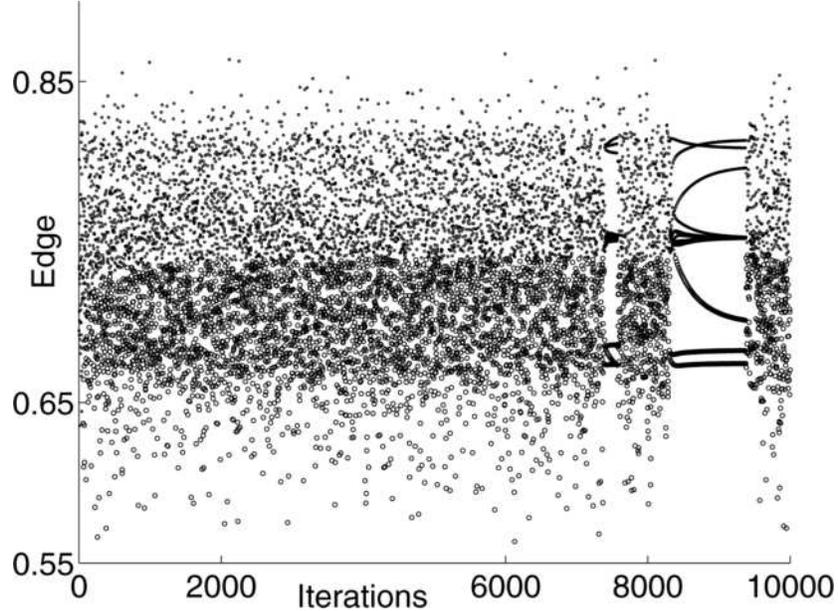

FIG. 3. *Value of the edge at each iteration $t$, for a run of AdaBoost using a $12 \times 25$ matrix* **M**. *Whenever $G$ increased from the current iteration to the following iteration, a small circle was plotted. Whenever $G$ decreased, a large circle was plotted. The fact that the larger circles are below the smaller circles is a direct result of Theorem 7.1. In fact, one can visually track the progress of $G$ using the boundary between the larger and smaller circles. For further explanation of the interesting dynamics in this plot, see [22].*

PROOF OF THEOREM 7.1. Using the expression $\alpha_t^{[A]} = \gamma_t = \tanh^{-1} r_t$ chosen by AdaBoost, the condition for $G$ to increase (or at least stay constant) is $G(\boldsymbol{\lambda}_t^{[A]}) \leq G(\boldsymbol{\lambda}_t^{[A]} + \alpha_t^{[A]} \mathbf{e}_{j_t}) = G(\boldsymbol{\lambda}_{t+1}^{[A]})$, which occurs if and only if

$$(s_t^{[A]} + \alpha_t^{[A]}) G(\boldsymbol{\lambda}_t^{[A]}) \leq (s_t^{[A]} + \alpha_t^{[A]}) G(\boldsymbol{\lambda}_{t+1}^{[A]}) = s_t^{[A]} G(\boldsymbol{\lambda}_t^{[A]}) + \int_0^{\alpha_t^{[A]}} \tanh u \, du,$$

that is,

$$G(\boldsymbol{\lambda}_t^{[A]}) \leq \left( \int_0^{\alpha_t^{[A]}} \tanh u \, du \right) \Big/ \alpha_t^{[A]} = \Upsilon(r_t),$$

where we have used the recursive equation (4.2) and the fact that $\alpha_t^{[A]}$ is a function of $r_t$. Thus, our statement is proved. $\square$

Hence, AdaBoost makes progress (measured by $G$) if and only if it takes a sufficiently large step. Figure 3 illustrates this point.

7.1. *Cyclic AdaBoost and the smooth margin.* It has been shown that AdaBoost's weight vectors $(\mathbf{d}_1, \mathbf{d}_2, \ldots)$ may converge to a stable periodic cycle [22]. In fact, the existence of these periodic cycles has already been an important tool for proving convergence properties of AdaBoost in the optimal case; thus far, they have provided the only nontrivial cases in which AdaBoost's convergence can be completely understood. Additionally, they have been used to show that AdaBoost may converge to a solution with margin significantly below maximum, even in the optimal case. This mysterious and beautiful cyclic behavior for AdaBoost often seems to occur when the number of training examples is small, although it has been observed in larger cases as well. Since this cycling phenomenon has proven so useful, we extend our earlier work [22] in this section.

While Algorithms 1 and 2 make progress with respect to $G$ at every iteration, we show that almost the opposite is true for AdaBoost when cycling occurs. Namely, we show that AdaBoost cannot increase $G$ at every iteration except under very special circumstances. For this theorem, we assume that AdaBoost is in the process of converging to a cycle, and not necessarily on the cycle itself. The edge values on the cycle are denoted $r_1^{\text{cyc}}, \ldots, r_T^{\text{cyc}}$, where the cycle has length $T$. (E.g., an edge close to $r_1^{\text{cyc}}$ is followed by an edge close to $r_2^{\text{cyc}}$, an edge close to $r_{T-1}^{\text{cyc}}$ is followed by an edge close to $r_T^{\text{cyc}}$, which is followed by an edge close to $r_1^{\text{cyc}}$. Note that there are cases where the limiting edge values $r_1^{\text{cyc}}, \ldots, r_T^{\text{cyc}}$ can be analytically determined from AdaBoost's dynamical formulas [22]. For our theorem, we do not need to assume these values are known, only that they exist.)

THEOREM 7.2 (Cyclic AdaBoost and the smooth margin). *Assume AdaBoost is converging to a cycle of $T$ iterations. Then one of the following conditions must be obeyed:*

1. *the value of $G$ decreases an infinite number of times, or*
2. *the edge values in the cycle $r_1^{\text{cyc}}, \ldots, r_T^{\text{cyc}}$ are equal (i.e., $r_1^{\text{cyc}} = \cdots = r_T^{\text{cyc}} = r$ and thus $r_t \to r$), and $G(\boldsymbol{\lambda}_t^{[A]}) \to \Upsilon(r)$ as $t \to \infty$.*

Thus, the value of $G$ cannot be strictly increasing except in this very special case where AdaBoost's edges, and thus its step sizes, are constant. This is in contrast to our new algorithms, which make significant progress toward increasing $G$ at each iteration. The proof of Theorem 7.2 can be found in Section 10.

Note that some important previously studied cases fall under the exceptional case 2 of Theorem 7.2 [22]. Hence we now look into case 2 further. In case 2, the value of $G$ is nondecreasing, and the values of $r_t^{\text{cyc}}$ are identical. Let us sort the training examples. Within a cycle, for training example $i$, either $d_{t,i} = 0\ \forall t$ or $d_{t,i} > 0\ \forall t$. The examples $i$ such that $d_{t,i} > 0\ \forall t$ are support vectors by definition. It can be shown that the support vectors also attain the same (minimum) margin [22]. It turns out that the support vectors have a nice property in this case, namely, they are treated equally by the weak learning algorithm in the following sense:

THEOREM 7.3 (Cyclic AdaBoost and the smooth margin—exceptional case). *Assume AdaBoost is within a cycle. If all edges in a cycle are the same, that is, $r_t = r \ \forall t$, then all support vectors are misclassified by the same number of weak classifiers within the cycle.*

PROOF. Consider support vectors $i$ and $i'$. Since they are support vectors, they must obey the cycle condition derived from AdaBoost's dynamical equations [22, 23], namely: $\prod_{t=1}^{T}(1 + M_{ij_t}r) = 1$ and $\prod_{t=1}^{T}(1 + M_{i'j_t}r) = 1$. Here we have assumed AdaBoost started on the cycle at iteration 1 without loss of generality. Define $\tau_i := |\{t : 1 \leq t \leq T, M_{ij_t} = 1\}|$. Here, $\tau_i$ represents the number of times example $i$ is correctly classified during one cycle, $1 \leq \tau_i \leq T$.

$$1 = \prod_{t=1}^{T}(1 + M_{ij_t}r) = (1+r)^{\tau_i}(1-r)^{T-\tau_i} = (1+r)^{\tau_{i'}}(1-r)^{T-\tau_{i'}}.$$

Hence, $\tau_i = \tau_{i'}$. Thus, example $i$ is classified correctly the same number of times that $i'$ is classified correctly. Since the choice of $i$ and $i'$ was arbitrary, this holds for all support vectors. □

This theorem shows that a stronger equivalence between support vectors exists here; not only do the support vectors achieve the same margin, but they are all "viewed" similarly by the weak learning algorithm, in that they are misclassified the same proportion of the time. As we have found no substantial correlation between the number of support vectors, the number of iterations in the cycle, and the number of rows or columns of $\mathbf{M}$, this result is somewhat surprising, especially since weak classifiers may appear more than once per cycle, so the number of weak classifiers is not even directly related to the number of iterations in a cycle.

Another observation is that even if the value of $G$ is nondecreasing for all iterations in the cycle (i.e, the exceptional case we have just discussed), AdaBoost may not converge to a maximum margin solution, as shown by an example analyzed in earlier work [22].

7.2. *Convergence of AdaBoost with bounded edges.* We will now give the direct relationship between edge values and margin values promised earlier. A special case of this result yields a proof that Rätsch and Warmuth's [17] bound on the margin achieved by AdaBoost is tight. This fixes the "gap in

BOOSTING AND THE SMOOTH MARGIN                    25

theory" used as the motivation for the development of AdaBoost$^*$. We will assume that throughout the run of AdaBoost, our weak classifiers always have edges within a small interval $[\bar{\rho}, \bar{\rho} + \sigma]$ where $\bar{\rho} \geq \rho$. As $\bar{\rho} \to \rho$ and $\sigma \to 0$ we approach the most extreme nonoptimal case. The justification for allowing a range of possible edge values is practical rather than theoretical; a weak learning algorithm will probably not be able to achieve an edge of exactly $\bar{\rho}$ at every iteration since the number of training examples is finite, and since the edge is a combinatorial quantity. Thus, we assume only that the edge is within a given interval rather than an exact value. Later we will give an example to show that we can force this interval to be arbitrarily small as long as the number of training examples is large enough.

THEOREM 7.4 (Convergence of AdaBoost with bounded edges). *Assume that for each $t$, AdaBoost's weak learning algorithm achieves an edge $r_t$ such that $r_t \in [\bar{\rho}, \bar{\rho} + \sigma]$ for some $\rho \leq \bar{\rho} < 1$ and for some $\sigma > 0$. Then,*

$$\limsup_{t \to \infty} g_t^{[A]} \leq \Upsilon(\bar{\rho} + \sigma)$$

*and*

$$\liminf_{t \to \infty} g_t^{[A]} \geq \Upsilon(\bar{\rho}).$$

*For the special case $\lim_{t \to \infty} r_t = \rho$, this implies*

$$\lim_{t \to \infty} g_t^{[A]} = \lim_{t \to \infty} \mu(\boldsymbol{\lambda}_t^{[A]}) = \Upsilon(\rho).$$

This result gives an explicit small range for the margin $\mu(\boldsymbol{\lambda}_t^{[A]})$, since from (3.2) and $\lim_{t \to \infty} \|\boldsymbol{\lambda}_t^{[A]}\|_1 \to \infty$, we have $\lim_{t \to \infty} (g_t^{[A]} - \mu(\boldsymbol{\lambda}_t^{[A]})) = 0$. (The statement $\lim_{t \to \infty} \|\boldsymbol{\lambda}_t^{[A]}\|_1 \to \infty$ always occurs for AdaBoost in the separable case since the edge is bounded above zero.) The special case $\lim_{t \to \infty} r_t = \rho$ shows the tightness of the bound of Rätsch and Warmuth [17] (see [15] for the proof). Their result, which we summarize only for AdaBoost rather than for the slightly more general AdaBoost$_\varrho$, states that $\liminf_{t \to \infty} \mu(\boldsymbol{\lambda}_t^{[A]}) \geq \Upsilon(r_{\inf})$, where $r_{\inf} = \inf_t r_t$. (The statement of their theorem seems to assume the existence of a combined hypothesis and limiting margin, but we believe these strong assumptions are not necessary, and that their proof of the lower bound holds without these assumptions.) Theorem 7.4 gives bounds from both above and below, so we now have a much more explicit convergence property of the margin. The proof can be found in Section 10.

Our next result is that Theorem 7.4 can be realized even for arbitrarily small interval size $\sigma$. In other words, AdaBoost can achieve any margin with arbitrarily high accuracy; that is, for a given margin value and precision, we can construct a training set and weak learning algorithm where AdaBoost attains that margin with that precision.

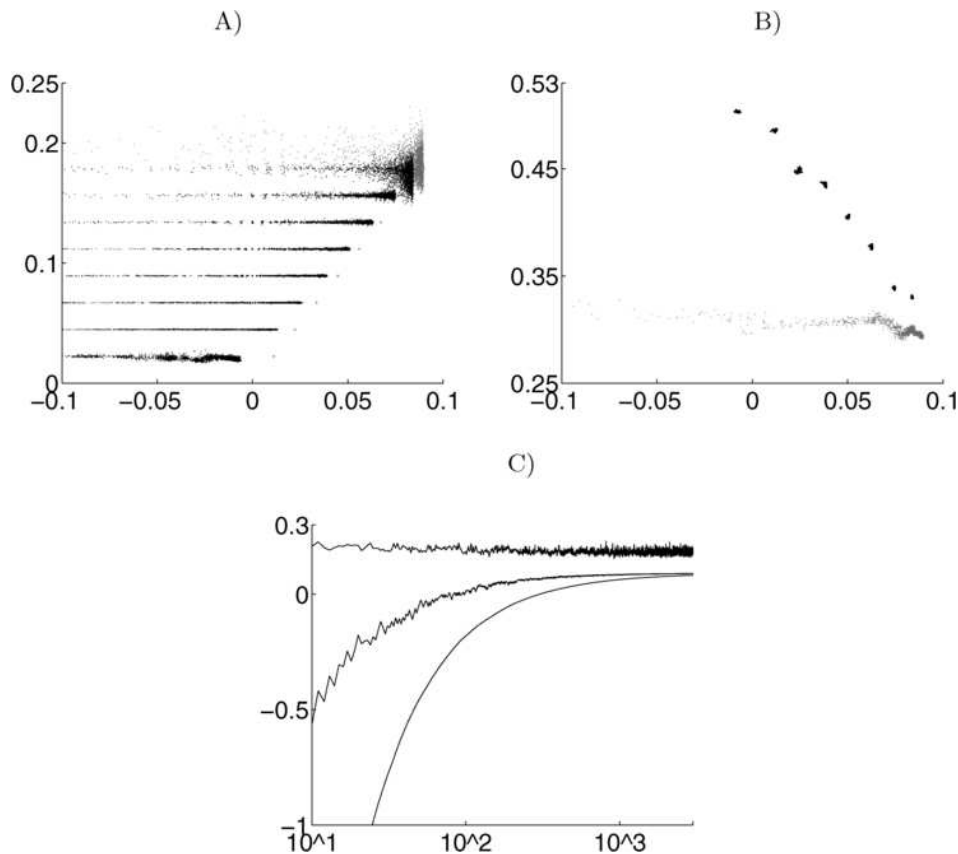

Fig. 4. *AdaBoost's probability of error on test data decreases as the margin increases. We computed nine trials, namely, eight trials of nonoptimal AdaBoost, $\ell = 1, \ldots, 8$, and one trial of optimal AdaBoost (denoted via $\ell = 0$). For each nonoptimal trial $\ell$, a goal edge value $\overline{r_\ell}$ was manually prespecified. For 3,000 iterations of each trial, we stored the edge values $r_{\ell,t}$ and margins $\mu_{\ell,t}$ on the training set, along with the probability of error on a randomly chosen test set $e_{\ell,t}$. A—edge versus margin. In each of the nine trials, we plot $(\mu_{\ell,t}, r_{\ell,t})$ for iterations $t$ that fall within the plot domain. Later iterations tend to give points nearer to the right in the plot. Additionally, dots have been placed at the points $(\Upsilon(\overline{r_\ell}), \overline{r_\ell})$ for $\ell = 1, \ldots, 8$. By Theorem 7.4, the asymptotic margin value for trial $\ell$ should be approximately $\Upsilon(\overline{r_\ell})$. Thus, AdaBoost's margins $\mu_{\ell,t}$ are converging to the prespecified margins $\Upsilon(\overline{r_\ell})$. B—probability of error versus margins. The lower scattered curve represents optimal AdaBoost; for optimal AdaBoost, we have plotted all $(\mu_{0,t}, e_{0,t})$ pairs falling within the plot domain. For clarity, we plot only the last 250 iterations for each nonoptimal trial, that is, for trial $\ell$, there is a clump of 250 points $(\mu_{\ell,t}, e_{\ell,t})$ with margin values $\mu_{\ell,t} \approx \Upsilon(\overline{r_\ell})$. This plot shows that the probability of error decreases as the prespecified margin increases. C—edges $r_{0,t}$ (top curve), margins $r_{0,t}$ (middle curve) and smooth margins (lower curve) versus number of iterations $t$ for only the optimal AdaBoost trial.*

THEOREM 7.5 (Bound of Theorem 7.4 is nonvacuous). *Say we are given $0 < \bar{\rho} < 1$ and $\sigma > 0$ arbitrarily small. Then there is some matrix $\mathbf{M}$ for which nonoptimal AdaBoost may choose an infinite sequence of weak classifiers with edge values in the interval $[\bar{\rho}, \bar{\rho} + \sigma]$. Additionally for this matrix $\mathbf{M}$, we have $\bar{\rho} \geq \rho$* (*where $\rho$ is the maximum margin for $\mathbf{M}$*).

The proof, in Section 10, is by explicit construction, in which the number of examples and weak classifiers increases as more precise bounds are required, that is, as the precision width parameter $\sigma$ decreases.

Let us see Theorem 7.4 in action. Now that one can more or less predetermine the value of AdaBoost's margin simply by choosing the edge values to be within a small range, one might again consider the important question of whether AdaBoost's asymptotic margin matters for generalization. To study this empirically, we use AdaBoost only, several times on the same data set with the same set of weak classifiers. Our results show that the choice of edge value (and thus the asymptotic margin) does have a dramatic effect on the test error. Artificial test data for Figure 4 was designed as follows: 300 examples were constructed randomly such that each $\mathbf{x}_i$ lies on a corner of the hypercube $\{-1, 1\}^{800}$. The labels are: $y_i = \text{sign}(\sum_{k=1}^{51} \mathbf{x}_i(k))$, where $\mathbf{x}_i(k)$ indicates the $k$th component of $\mathbf{x}_i$. For $j = 1, \ldots, 800$, the $j$th weak classifier is $h_j(\mathbf{x}) = \mathbf{x}(j)$, thus $M_{ij} = y_i \mathbf{x}_i(j)$. For $801 \leq j \leq 1600$, $h_j = -h_{(j-800)}$. There were 10,000 identically distributed randomly generated examples used for testing. The hypothesis space must be the same for each trial as a control; we purposely did not restrict the space via regularization (e.g., norm regulation, early stopping, or pruning). Hence we have a controlled experiment where only the choice of weak classifier is different, and this directly determines the margin via Theorem 7.4. AdaBoost was run nine times on this dataset, each time for $t_{\max} = 3{,}000$ iterations, the first time with standard optimal-case AdaBoost, and eight times with nonoptimal AdaBoost. For each nonoptimal trial, we selected a "goal" edge value $r_{\text{goal}}$ (the eight goal edge values were equally spaced). The weak learning algorithm chooses the closest possible edge to that goal. In this way, AdaBoost's margin is close to $\Upsilon(r_{\text{goal}})$. The results are shown in Figure 4B, which shows test error versus margins for the asymptotic regime of optimal AdaBoost (lower scattered curve) and the last 250 iterations for each nonoptimal trial (the eight clumps, each containing 250 points). It is very clear that as the margin increases, the probability of error decreases, and optimal AdaBoost has the lowest probability of error.

Note that the *asymptotic* margin is not the whole story; optimal AdaBoost yields a lower probability of error even before the asymptotic regime was reached. Thus, it is the degree of "optimal-ness" of the weak learning algorithm (directly controlling the asymptotic margin) that is inversely correlated with the probability of error for AdaBoost.

Now that we have finished describing the results, we move on to the proofs.

**8. Proof of Proposition 3.1.** To show property 1 given assumptions on $\mathbf{M}$, we will compute an arbitrary element of the Hessian $\mathbf{H}$,

$$H_{kj} = \frac{\partial^2 G(\boldsymbol{\lambda})}{\partial \lambda_k \, \partial \lambda_j} = -\frac{\frac{\partial^2 F(\boldsymbol{\lambda})}{\partial \lambda_k \, \partial \lambda_j}}{F(\boldsymbol{\lambda})\|\boldsymbol{\lambda}\|_1} + \frac{\frac{\partial F(\boldsymbol{\lambda})}{\partial \lambda_j}}{F(\boldsymbol{\lambda})\|\boldsymbol{\lambda}\|_1^2} + \frac{\frac{\partial F(\boldsymbol{\lambda})}{\partial \lambda_j} \frac{\partial F(\boldsymbol{\lambda})}{\partial \lambda_k}}{F(\boldsymbol{\lambda})^2 \|\boldsymbol{\lambda}\|_1}$$

$$+ \frac{\frac{\partial F(\boldsymbol{\lambda})}{\partial \lambda_k}}{F(\boldsymbol{\lambda})\|\boldsymbol{\lambda}\|_1^2} - \frac{2 \ln F(\boldsymbol{\lambda})}{\|\boldsymbol{\lambda}\|_1^3}.$$

For $G$ to be concave, we need $\mathbf{w}^T \mathbf{H} \mathbf{w} \leq 0$ for all vectors $\mathbf{w}$. We are considering the case where $\mathbf{w}$ obeys $\sum_j w_j = 0$ so we are considering only directions in which $\|\boldsymbol{\lambda}\|_1$ does not change. Thus, we are showing that $G$ is concave on every "shell." Note that $\sum_{j,k} w_j w_k \frac{\partial F(\boldsymbol{\lambda})}{\partial \lambda_j} = (\sum_j w_j \frac{\partial F(\boldsymbol{\lambda})}{\partial \lambda_j})(\sum_k w_k) = 0$, and thus

$$\sum_{j,k} w_j w_k H_{kj}$$

(8.1)
$$= \frac{-1}{F(\boldsymbol{\lambda})\|\boldsymbol{\lambda}\|_1} \sum_{j,k} w_j w_k \frac{\partial^2 F(\boldsymbol{\lambda})}{\partial \lambda_k \, \partial \lambda_j} + 0$$
$$+ \frac{1}{F(\boldsymbol{\lambda})^2 \|\boldsymbol{\lambda}\|_1} \left[\sum_j w_j \frac{\partial F(\boldsymbol{\lambda})}{\partial \lambda_j}\right]^2 + 0 + 0$$
$$= \frac{1}{F(\boldsymbol{\lambda})^2 \|\boldsymbol{\lambda}\|_1} \left[\left(-\sum_{i=1}^m (\mathbf{M}\mathbf{w})_i^2 e^{-(\mathbf{M}\boldsymbol{\lambda})_i}\right)\left(\sum_{i=1}^m e^{-(\mathbf{M}\boldsymbol{\lambda})_i}\right) \right.$$
$$\left. + \left[\sum_{i=1}^m (\mathbf{M}\mathbf{w})_i e^{-(\mathbf{M}\boldsymbol{\lambda})_i}\right]^2\right].$$

Let the vectors $\boldsymbol{\Psi}_1$ and $\boldsymbol{\Psi}_2$ be defined as $\boldsymbol{\Psi}_{1,i} := (\mathbf{M}\mathbf{w})_i e^{-(\mathbf{M}\boldsymbol{\lambda})_i/2}$ and $\boldsymbol{\Psi}_{2,i} := e^{-(\mathbf{M}\boldsymbol{\lambda})_i/2}$. The Cauchy–Schwarz inequality applied to $\boldsymbol{\Psi}_1$ and $\boldsymbol{\Psi}_2$ gives

$$-\left(\sum_{i=1}^m \boldsymbol{\Psi}_{1,i}^2\right)\left(\sum_{i=1}^m \boldsymbol{\Psi}_{2,i}^2\right) + \left(\sum_{i=1}^m \boldsymbol{\Psi}_{1,i}\boldsymbol{\Psi}_{2,i}\right)^2 \leq 0.$$

Since this expression is identical to the one bracketed in (8.1), $\sum_{j,k} w_j w_k H_{kj} \leq 0$, and thus we have shown that the function $G(\boldsymbol{\lambda})$ is concave on each shell, but not strictly. Equality in the Cauchy–Schwarz equation is achieved only when $\boldsymbol{\Psi}_1$ is parallel to $\boldsymbol{\Psi}_2$, that is, when $(\mathbf{M}\mathbf{w})_i$ does not depend on $i$. There are some matrices where such a $\mathbf{w}$ exists, for example, the matrix

$$\mathbf{M} = \begin{pmatrix} -1 & 1 & 1 & 1 \\ 1 & -1 & 1 & -1 \\ 1 & 1 & -1 & -1 \end{pmatrix}$$

with vector $\mathbf{w} = (-\frac{1}{2}c, c, c, -\frac{3}{2}c)$, where $c \in \mathbb{R}$. Here, $(\mathbf{Mw})_i = c$ for all $i$. We have shown that the function $G$ is concave for each "shell," but not necessarily strictly concave. (One can find out whether $G$ is concave on each shell for a particular matrix $\mathbf{M}$ by solving $\mathbf{Mw} = c\mathbf{1}$ subject to $\sum_j w_j = 0$, which can be added as a row.) We have now finished the proof of property 1.

To show property 2, we compute the derivative in the radial direction, $dG(\boldsymbol{\lambda}(1+a))/da|_{a=0}$, and show that it is positive. We find, using the notation $d_i := e^{-(\mathbf{M}\boldsymbol{\lambda})_i}/F(\boldsymbol{\lambda})$,

$$\begin{aligned}
\frac{dG(\boldsymbol{\lambda}(1+a))}{da}\bigg|_{a=0} &= \frac{1}{\|\boldsymbol{\lambda}\|_1}\left[\sum_{i=1}^m d_i(\mathbf{M}\boldsymbol{\lambda})_i + \ln F(\boldsymbol{\lambda})\right] \\
&\geq \frac{1}{\|\boldsymbol{\lambda}\|_1}\left[\left(\sum_{i=1}^m d_i\right)\min_{\tilde{i}}(\mathbf{M}\boldsymbol{\lambda})_{\tilde{i}} + \ln\sum_{i=1}^m e^{-(\mathbf{M}\boldsymbol{\lambda})_i}\right] \\
&> \frac{1}{\|\boldsymbol{\lambda}\|_1}\left[\min_{\tilde{i}}(\mathbf{M}\boldsymbol{\lambda})_{\tilde{i}} + \ln e^{-\min_{\tilde{i}}(\mathbf{M}\boldsymbol{\lambda})_{\tilde{i}}}\right] = 0.
\end{aligned}$$

The very last inequality follows since from our $m > 1$ terms, we took only one term, and also since $\sum_i d_i = 1$.

**9. Convergence proofs.** Before we state the proofs, we must continue our simplification of the recursive equations. From the recursive equation for $G$, namely (4.2) applied to Algorithm 1,

(9.1)
$$\begin{aligned}
s_{t+1}^{[1]}g_{t+1}^{[1]} - s_t^{[1]}g_t^{[1]} &= \ln\left(\frac{\cosh\gamma_t}{\cosh(\gamma_t - \alpha_t^{[1]})}\right) = \frac{1}{2}\ln\left(\frac{1 - \tanh^2(\gamma_t - \alpha_t^{[1]})}{1 - \tanh^2\gamma_t}\right) \\
&= \frac{1}{2}\ln\frac{[1 - \tanh(\gamma_t - \alpha_t^{[1]})][1 + \tanh(\gamma_t - \alpha_t^{[1]})]}{(1 - \tanh\gamma_t)(1 + \tanh\gamma_t)} \\
&= \frac{1}{2}\ln\frac{(1 - g_{t+1}^{[1]})(1 + g_{t+1}^{[1]})}{(1 - r_t)(1 + r_t)} = \alpha_t^{[1]} + \ln\left(\frac{1 + g_{t+1}^{[1]}}{1 + r_t}\right).
\end{aligned}$$

Here we have used both (4.3) and (4.4). We perform an analogous simplification for Theorem 5.2. Starting from (4.2) and applying (4.5) and (4.6),

(9.2)
$$\begin{aligned}
s_{t+1}^{[2]}g_{t+1}^{[2]} - s_t^{[2]}g_t^{[2]} &= \frac{1}{2}\ln\frac{(1 - g_t^{[2]})(1 + g_t^{[2]})}{(1 - r_t)(1 + r_t)} \\
&= \alpha_t^{[2]} + \ln\left(\frac{1 + g_t^{[2]}}{1 + r_t}\right).
\end{aligned}$$

We will use equations (9.1) and (9.2) to help us with the proofs.

PROOF OF LEMMA 5.1. We start with Algorithm 2. First, we note that since the function tanh is concave on $\mathbb{R}_+$, we can lower bound tanh on an interval $(a,b) \subset (0,\infty)$ by the line connecting the points $(a, \tanh(a))$ and $(b, \tanh(b))$. Thus,

$$\int_{\gamma_t - \alpha_t^{[2]}}^{\gamma_t} \tanh u \, du \geq \tfrac{1}{2} \alpha_t^{[2]} [\tanh \gamma_t + \tanh(\gamma_t - \alpha_t^{[2]})]$$

(9.3)
$$= \tfrac{1}{2} \alpha_t^{[2]} (r_t + g_t),$$

where the last equality is from (4.5). Combining (9.3) with (4.2) yields

$$s_{t+1}^{[2]} g_{t+1}^{[2]} - s_t g_t \geq \tfrac{1}{2} \alpha_t^{[2]} (r_t + g_t),$$

$$s_{t+1}^{[2]} (g_{t+1}^{[2]} - g_t) + \alpha_t^{[2]} g_t \geq \tfrac{1}{2} \alpha_t^{[2]} (r_t + g_t),$$

$$g_{t+1}^{[2]} - g_t \geq \frac{\alpha_t^{[2]} (r_t - g_t)}{2 s_{t+1}^{[2]}}.$$

Thus, the statement of the lemma holds for Algorithm 2. By definition, $g_{t+1}^{[1]}$ is the maximum value of $G(\boldsymbol{\lambda}_t + \alpha \mathbf{e}_{j_t})$, so $g_{t+1}^{[1]} \geq g_{t+1}^{[2]}$. By (4.4) and (4.6), we know $\alpha_t^{[1]} \leq \alpha_t^{[2]}$. Because $\alpha/(s+\alpha) = 1 - s/(\alpha + s)$ increases with $\alpha$,

$$g_{t+1}^{[1]} - g_t \geq g_{t+1}^{[2]} - g_t \geq \left(\frac{\alpha_t^{[2]}}{s_{t+1}^{[2]}}\right) \frac{(r_t - g_t)}{2} \geq \left(\frac{\alpha_t^{[1]}}{s_{t+1}^{[1]}}\right) \frac{(r_t - g_t)}{2}.$$

Thus, we have completed the proof of Lemma 5.1. $\square$

PROOF OF LEMMA 5.2. The proof holds for both algorithms, so we have dropped the superscripts. There are two possibilities; either $\lim_{t\to\infty} s_t = \infty$ or $\lim_{t\to\infty} s_t < \infty$. We handle these cases separately, starting with the case $\lim_{t\to\infty} s_t = \infty$. From (9.1) and (9.2), and recalling that $g_t \leq g_{t+1} \leq \rho \leq r_t$ we know

$$s_{t+1} g_{t+1} - s_t g_t \geq \alpha_t + \ln \frac{1 + g_t}{1 + r_t},$$

so that

$$\alpha_t (1 - \rho) \leq \alpha_t (1 - g_{t+1}) \leq s_t (g_{t+1} - g_t) + \ln \frac{1 + r_t}{1 + g_t}.$$

We denote by $\tilde{1}$ the first iteration where $G$ is positive, so $g_{\tilde{1}} > 0$. Dividing by $(1-\rho) s_t$, recalling that $r_t < 1$ and $g_{\tilde{1}} \leq g_t$,

$$\frac{\alpha_t}{s_{t+1}} \leq \frac{\alpha_t}{s_t} \leq \frac{g_{t+1} - g_t}{1 - \rho} + \frac{1}{1 - \rho} \frac{1}{s_t} \ln \frac{1 + r_t}{1 + g_t}$$

$$\leq \frac{g_{t+1} - g_t}{1 - \rho} + \frac{1}{1 - \rho} \frac{1}{s_t} \ln \frac{2}{1 + g_{\tilde{1}}}.$$

We will take the limit of both sides as $t \to \infty$. Since the values $g_t$ are monotonically increasing and are bounded by 1, $\lim_{t\to\infty}(g_{t+1} - g_t) = 0$. Hence, the first term vanishes in the limit. Since $\lim_{t\to\infty} s_t = \infty$, the second term also vanishes in the limit. Thus, the statement of the lemma holds when $s_t \to \infty$.

Now for the case where $\lim_{t\to\infty} s_t < \infty$, consider

$$\sum_{t=\tilde{1}}^{T} \frac{\alpha_t}{s_{t+1}} = \sum_{t=\tilde{1}}^{T} \frac{s_{t+1} - s_t}{s_{t+1}} = \sum_{t=\tilde{1}}^{T} \int_{s_t}^{s_{t+1}} \frac{1}{s_{t+1}} du$$

$$\leq \sum_{t=\tilde{1}}^{T} \int_{s_t}^{s_{t+1}} \frac{1}{u} du = \int_{s_{\tilde{1}}}^{s_{T+1}} \frac{1}{u} du = \ln \frac{s_{T+1}}{s_{\tilde{1}}}.$$

By our assumption that $\lim_{t\to\infty} s_t < \infty$, the above sequence is a bounded increasing sequence. Thus, $\sum_{t=\tilde{1}}^{\infty} \alpha_t/s_{t+1}$ converges. In particular, $\lim_{t\to\infty} \alpha_t/s_{t+1} = 0$. □

PROOF OF THEOREM 5.1. We choose to show convergence from the starting position $\boldsymbol{\lambda}_{\tilde{1}}$, where $\boldsymbol{\lambda}_{\tilde{1}}$ is the coefficient vector at the first iteration where $G$ is positive. This is the iteration where we switch from AdaBoost to our new iteration scheme; it suffices to show convergence from this point. For this proof, we drop the superscripts [1] and [2]; each step in the proof holds for both algorithms.

The values of $g_t$ constitute a nondecreasing sequence that is uniformly bounded by 1. Thus, a limit $g_\infty$ must exist, $g_\infty := \lim_{t\to\infty} g_t$. By (3.2), we know that $g_t \leq \rho$ for all $t$. Thus, $g_\infty \leq \rho$. Let us suppose that $g_\infty < \rho$, that is, that $\rho - g_\infty \neq 0$. (We will show this assumption is not true by contradiction.)

From Lemma 5.2, there exists a time $t_0 \in \mathbb{N}$ such that, for all times $t \geq t_0$, we have $\alpha_t/s_{t+1} \leq 1/2$, or equivalently, $\alpha_t \leq s_{t+1}/2$, and thus $s_t = s_{t+1} - \alpha_t \geq s_{t+1}/2$, so that

$$(9.4) \qquad \frac{\alpha_t}{s_t} \leq \frac{2\alpha_t}{s_{t+1}} \qquad \text{for } t \geq t_0.$$

From Lemma 5.1, since $g_t \leq g_\infty$ and $r_t \geq \rho$, we have

$$(\rho - g_\infty)\frac{\alpha_t}{2s_{t+1}} \leq \frac{\alpha_t}{s_{t+1}} \frac{(r_t - g_t)}{2} \leq g_{t+1} - g_t.$$

Thus, for all $T \in \mathbb{N}$,

$$(9.5) \qquad (\rho - g_\infty) \sum_{t=\tilde{1}}^{T} \frac{\alpha_t}{2s_{t+1}} \leq \sum_{t=\tilde{1}}^{T} (g_{t+1} - g_t) = g_{T+1} - g_{\tilde{1}} < 1.$$

Under our assumption $\rho - g_\infty \neq 0$, the inequality (9.5) implies that the series $\sum_{t=\tilde{1}}^{\infty} (\alpha_t/s_{t+1})$ converges. This, combined with (9.4), implies that the series $\sum_{t=\tilde{1}}^{\infty}(\alpha_t/s_t)$ converges, since its tail is majorized, term by term, by the tail of a converging series. Therefore, for all $T \in \mathbb{N}$, $T > 1$,

$$\infty > \sum_{t=\tilde{1}}^{\infty} \frac{\alpha_t}{s_t} \geq \sum_{t=\tilde{1}}^{T-1} \frac{\alpha_t}{s_t} = \sum_{t=\tilde{1}}^{T-1} \frac{s_{t+1} - s_t}{s_t} = \sum_{t=\tilde{1}}^{T-1} \int_{s_t}^{s_{t+1}} \frac{1}{s_t} du$$

$$\geq \sum_{t=\tilde{1}}^{T-1} \int_{s_t}^{s_{t+1}} \frac{1}{u} du = \int_{s_{\tilde{1}}}^{s_T} \frac{1}{u} du = \ln s_T - \ln s_{\tilde{1}}.$$

Therefore, the $s_t$ constitute a bounded, increasing sequence and must converge; define $s_\infty := \lim_{T \to \infty} s_T < \infty$. The convergence of the $s_t$ sequence implies that $\alpha_t = s_{t+1} - s_t$ must converge to zero: $\lim_{t \to \infty} \alpha_t = 0$. Finally, we use the fact that tanh is continuous and strictly increasing, together with (4.3) and (4.5), to derive

$$g_\infty = \lim_{t \to \infty} g_t = \liminf_{t \to \infty} g_t = \tanh\left[\liminf_{t \to \infty}(\gamma_t - \alpha_t)\right]$$

$$= \tanh\left[\liminf_{t \to \infty} \gamma_t - \lim_{t \to \infty} \alpha_t\right] = \tanh\left[\liminf_{t \to \infty} \gamma_t\right] = \liminf_{t \to \infty}[\tanh \gamma_t]$$

$$= \liminf_{t \to \infty} r_t \geq \rho.$$

This is a contradiction with the original assumption that $g_\infty < \rho$. It follows that we have proved that $g_\infty = \rho$, or $\lim_{t \to \infty}(\rho - g_t) = 0$. □

PROOF OF LEMMA 5.3. The proof works for both algorithms, so we leave off the superscripts. From (4.2),

(9.6) $$s_{t+1} g_{t+1} - s_t g_t = \ln \cosh \gamma_t - \ln \cosh(\gamma_t - \alpha_t).$$

Because $(1/2)e^\xi \leq 1/2(e^\xi + e^{-\xi}) = \cosh \xi \leq e^\xi$ for $\xi > 0$, we have $\xi - \ln 2 \leq \ln \cosh \xi \leq \xi$. Combining this with (9.6),

$$s_{t+1} g_{t+1} - s_t g_t \geq \gamma_t - \ln 2 - (\gamma_t - \alpha_t),$$

so

$$\alpha_t(1 - \rho) \leq \alpha_t(1 - g_{t+1}) \leq \ln 2 + s_t(g_{t+1} - g_t) \leq \ln 2 + \rho s_t.$$

The first and last inequalities of the last line use the fact that $G$ is positive and bounded by $\rho$, that is, $1 - \rho \leq 1 - g_{t+1}$ and $g_{t+1} - g_t \leq \rho$. Thus, dividing both sides by $(1 - \rho)$, we find the statement of the lemma. □

PROOF OF THEOREM 5.2. Again the superscripts have been removed since all statements are true for both algorithms. Define $\Delta G(\boldsymbol{\lambda}) := \rho - G(\boldsymbol{\lambda})$. Since (3.2) states that $g_t \leq \mu(\boldsymbol{\lambda}_t)$, we know $0 \leq \rho - \mu(\boldsymbol{\lambda}_t) \leq \rho - g_t = \Delta G(\boldsymbol{\lambda}_t)$, and thus we need only to control how fast $\Delta G(\boldsymbol{\lambda}_t) \to 0$ as $t \to \infty$. That is, if $g_t$ is within $\varepsilon$ of the maximum margin $\rho$, so is the margin $\mu(\boldsymbol{\lambda}_t)$. Starting from Lemma 5.1,

$$\rho - g_{t+1} \leq \rho - g_t - \frac{\alpha_t}{2s_{t+1}}(r_t - \rho + \rho - g_t),$$

thus

(9.7)
$$\Delta G(\boldsymbol{\lambda}_{t+1}) \leq \Delta G(\boldsymbol{\lambda}_t)\left[1 - \frac{\alpha_t}{2s_{t+1}}\right] - \frac{\alpha_t(r_t - \rho)}{2s_{t+1}}$$
$$\leq \Delta G(\boldsymbol{\lambda}_t)\left[1 - \frac{\alpha_t}{2s_{t+1}}\right] \leq \Delta G(\boldsymbol{\lambda}_{\tilde{1}})\prod_{\ell=\tilde{1}}^{t}\left[1 - \frac{\alpha_\ell}{2s_{\ell+1}}\right].$$

Here, the second inequality is due to the restriction $r_t \geq \rho$ and the fact that $\alpha_t > 0$. The last inequality of (9.7) is from the recursion. We stop the recursion at $\boldsymbol{\lambda}_{\tilde{1}}$, where $\boldsymbol{\lambda}_{\tilde{1}}$ is the coefficient vector at the first iteration where $G$ is positive. Before we continue, we upper bound the product in (9.7),

$$\prod_{\ell=\tilde{1}}^{t}\left[1 - \frac{\alpha_\ell}{2s_{\ell+1}}\right] = \prod_{\ell=\tilde{1}}^{t}\left[1 - \frac{1}{2}\frac{s_{\ell+1} - s_\ell}{s_{\ell+1}}\right] \leq \exp\left[-\frac{1}{2}\sum_{\ell=\tilde{1}}^{t}\frac{s_{\ell+1} - s_\ell}{s_{\ell+1}}\right]$$
$$\leq \exp\left[-\frac{1}{2}\sum_{\ell=\tilde{1}}^{t}\frac{s_{\ell+1} - s_\ell}{s_\ell + \rho/(1-\rho)s_\ell + \ln 2/(1-\rho)}\right]$$

(9.8)
$$= \exp\left[-\frac{1-\rho}{2}\sum_{\ell=\tilde{1}}^{t}\frac{s_{\ell+1} - s_\ell}{s_\ell + \ln 2}\right]$$
$$\leq \exp\left[-\frac{1-\rho}{2}\int_{s_{\tilde{1}}}^{s_{t+1}}\frac{dv}{v + \ln 2}\right]$$
$$= \left[\frac{s_{\tilde{1}} + \ln 2}{s_{t+1} + \ln 2}\right]^{(1-\rho)/2}.$$

Here, the first line holds since $1 - x \leq e^{-x}$ for all $x$, and the next line follows from our bound on the size of $\alpha_t$ in Lemma 5.3. Plugging back into (9.7), it follows that

$$\Delta G(\boldsymbol{\lambda}_t) \leq \Delta G(\boldsymbol{\lambda}_{\tilde{1}})\left[\frac{s_{\tilde{1}} + \ln 2}{s_t + \ln 2}\right]^{(1-\rho)/2},$$

or

(9.9)
$$s_t \leq s_t + \ln 2 \leq (s_{\tilde{1}} + \ln 2)\left[\frac{\Delta G(\boldsymbol{\lambda}_{\tilde{1}})}{\Delta G(\boldsymbol{\lambda}_t)}\right]^{2/(1-\rho)}.$$

On the other hand, we have (for Algorithm 2)

$$\alpha_t^{[2]} \geq \tanh \alpha_t^{[2]} = \tanh[\gamma_t - (\gamma_t - \alpha_t^{[2]})] = \frac{\tanh \gamma_t - \tanh(\gamma_t - \alpha_t^{[2]})}{1 - \tanh \gamma_t \tanh(\gamma_t - \alpha_t^{[2]})}$$

$$= \frac{r_t - g_t^{[2]}}{1 - r_t g_t^{[2]}} \geq \frac{\rho - g_t^{[2]}}{1 - \rho g_{\tilde{1}}} = \frac{\Delta G(\boldsymbol{\lambda}_t^{[2]})}{1 - \rho g_{\tilde{1}}} \geq \frac{\Delta G(\boldsymbol{\lambda}_{t+1}^{[2]})}{1 - \rho g_{\tilde{1}}}.$$

A similar calculation for Algorithm 1 holds. Thus, for both algorithms we have $\alpha_t \geq \Delta G(\boldsymbol{\lambda}_{t+1})/(1 - \rho g_{\tilde{1}})$ which implies

(9.10)
$$s_{t+1} = s_{\tilde{1}} + \sum_{\ell=\tilde{1}}^{t} \alpha_\ell \geq s_{\tilde{1}} + \sum_{\ell=\tilde{1}}^{t} \frac{\Delta G(\boldsymbol{\lambda}_{\ell+1})}{1 - \rho g_{\tilde{1}}}$$

$$\geq s_{\tilde{1}} + (t - \tilde{1} + 1)\frac{\Delta G(\boldsymbol{\lambda}_{t+1})}{1 - \rho g_{\tilde{1}}}.$$

Combining (9.9) with (9.10) leads to

$$t - \tilde{1} \leq \frac{(1 - \rho g_{\tilde{1}})s_t}{\Delta G(\boldsymbol{\lambda}_t)} \leq \frac{(1 - \rho g_{\tilde{1}})(s_{\tilde{1}} + \ln 2)[\Delta G(\boldsymbol{\lambda}_{\tilde{1}})]^{2/(1-\rho)}}{[\Delta G(\boldsymbol{\lambda}_t)]^{1+[2/(1-\rho)]}}$$

$$\leq \frac{s_{\tilde{1}} + \ln 2}{[\Delta G(\boldsymbol{\lambda}_t)]^{(3-\rho)/(1-\rho)}},$$

where we have used that $(1 - \rho g_{\tilde{1}}) \leq 1$, $\Delta G(\boldsymbol{\lambda}_{\tilde{1}}) \leq 1$. This means that $\Delta G(\boldsymbol{\lambda}_t) \geq \varepsilon$ is possible only if $t \leq \tilde{1} + (s_{\tilde{1}} + \ln 2)\varepsilon^{-(3-\rho)/(1-\rho)}$. Therefore, if $t$ exceeds $\tilde{1} + (s_{\tilde{1}} + \ln 2)\varepsilon^{-(3-\rho)/(1-\rho)}$, it follows that $\Delta G(\boldsymbol{\lambda}_t) < \varepsilon$. This concludes the proof of Theorem 5.2. □

PROOF OF LEMMA 5.4. We show that there is a $T_\nu$ such that after iteration $T_\nu$, $s_t^\nu(\rho - g_t)$ is a decreasing sequence,

$$s_{t+1}^\nu(\rho - g_{t+1}) \leq s_t^\nu(\rho - g_t) \qquad \text{for } t \geq T_\nu.$$

In this way, the value of $C_\nu$ will be determined by

$$C_\nu = \max_{t \in \{\tilde{1}, \ldots, T_\nu\}} s_t^\nu(\rho - g_t).$$

Let us examine our sufficient condition more closely. Using Lemma 5.1 we have, for arbitrary $t$,

$$s_t^\nu(\rho - g_t) - s_{t+1}^\nu(\rho - g_{t+1}) = (s_t^\nu - s_{t+1}^\nu)(\rho - g_t) + s_{t+1}^\nu(g_{t+1} - g_t)$$

(9.11)
$$\geq (s_t^\nu - s_{t+1}^\nu)(\rho - g_t) + s_{t+1}^\nu \frac{\alpha_t(r_t - g_t)}{2s_{t+1}}$$

$$\geq (s_t^\nu - s_{t+1}^\nu)(\rho - g_t) + s_{t+1}^\nu \frac{\alpha_t(\rho - g_t)}{2s_{t+1}}$$

$$= (\rho - g_t)\left[s_t^\nu - s_{t+1}^\nu + \frac{1}{2}s_{t+1}^{\nu-1}(s_{t+1} - s_t)\right].$$

Thus, it is sufficient to show that the bracketed term in (9.11) is positive for all sufficiently large $t$.

From Lemma 5.2, we know that for an arbitrary choice of $\varepsilon > 0$, there exists an iteration $t_\varepsilon$ such that for all $t \geq t_\varepsilon$, we have $\alpha_t/s_{t+1} \leq \varepsilon$. We will choose $\varepsilon = \varepsilon_\nu := 1 - (2\nu)^{1/(1-\nu)}$, for reasons that will become clear later. The corresponding iteration $t_{\varepsilon_\nu}$ will be the $T_\nu$ we are looking for. For $t \geq T_\nu$, we thus have

$$s_t = s_{t+1} - \alpha_t = s_{t+1}(1 - \tau_t) \qquad \text{for some } 0 \leq \tau_t \leq \varepsilon_\nu.$$

Using this to rewrite the bracketed terms of (9.11) yields

$$s_t^\nu - s_{t+1}^\nu + \tfrac{1}{2}s_{t+1}^{\nu-1}(s_{t+1} - s_t) = s_{t+1}^\nu[(1 - \tau_t)^\nu - 1 + \tfrac{1}{2}\tau_t],$$

so that the original claim will follow if we can prove that

$$f(\tau) := (1 - \tau)^\nu - 1 + \tfrac{1}{2}\tau \geq 0 \text{ for } \tau \in [0, \varepsilon_\nu].$$

We have $f(0) = 0$, and also, $f'(\tau) = 1/2 - \nu(1 - \tau)^{\nu-1}$. Because $1/2 < \nu < 1$, $f'(\tau)$ is a decreasing function of $\tau$; by the choice of $\varepsilon_\nu$, $f'(\varepsilon_\nu) = 0$, so that $f'(\tau) \geq 0$ for $\tau \in [0, \varepsilon_\nu]$. Hence $f(\tau)$ is an increasing function, which is positive for $\tau \in [0, \varepsilon_\nu]$. We have finished the proof of the lemma. $\square$

PROOF OF THEOREM 5.3. Most of the work has already been done in the proof of Theorem 5.2. By (9.10), we have $t - \tilde{1} \leq (1 - \rho g_{\tilde{1}})(\rho - g_t)^{-1}(s_t - s_{\tilde{1}})$. Combining this with Lemma 5.4 leads to

$$t - \tilde{1} \leq (1 - \rho g_{\tilde{1}})C_\nu^{1/\nu}(\rho - g_t)^{-(1+1/\nu)}.$$

For $\delta > 0$, we pick $\nu = \nu_\delta := 1/(2 + \delta) < 1/2$, and we can rewrite the last inequality as

$$(\rho - g_t)^{3+\delta} \leq (1 - \rho g_{\tilde{1}})C_{\nu_\delta}^{2+\delta}(t - \tilde{1})^{-1},$$

or more concisely, $\rho - g_t \leq C_\delta(t - \tilde{1})^{-1/(3+\delta)}$, where

$$C_\delta = (1 - \rho g_{\tilde{1}})^{1/(3+\delta)}C_{\nu_\delta}^{(2+\delta)/(3+\delta)}.$$

It follows that $\rho - \mu(\boldsymbol{\lambda}_t) \leq \rho - g_t < \varepsilon$ whenever $t - \tilde{1} > (C_\delta \varepsilon^{-1})^{(3+\delta)}$, which completes the proof of Theorem 5.3. $\square$

PROOF OF THEOREM 5.4. We use the notation $g_t = g_t^{[2]}$, $s_t = s_t^{[2]}$, and so forth, since we are using only Algorithm 2. Since $r_t = \rho$ for all $t$, we automatically have

$$(9.12) \quad s_{t+1} = s_t + \alpha_t = s_t + \frac{1}{2}\ln\left(\frac{1+\rho}{1-\rho}\frac{1-g_t}{1+g_t}\right),$$

and from (9.2),

$$(9.13) \quad s_{t+1}g_{t+1} = s_t g_t + \frac{1}{2}\ln\left(\frac{1+g_t}{1+\rho}\frac{1-g_t}{1-\rho}\right).$$

We will simplify these equations a number of times. For this proof only, we use the notation $x_t := \Delta G(\boldsymbol{\lambda}_t^{[2]}) := \rho - g_t$ to rewrite the quantities

$$\frac{1+g_t}{1+\rho} = 1 - \frac{x_t}{1+\rho} \quad \text{and} \quad \frac{1-g_t}{1-\rho} = 1 + \frac{x_t}{1-\rho}.$$

Using this notation, we update (9.12) and (9.13),

$$(9.14) \quad s_{t+1} = s_t + \frac{1}{2}\ln\left(1 + \frac{x_t}{1-\rho}\right) - \frac{1}{2}\ln\left(1 - \frac{x_t}{1+\rho}\right),$$

$$(9.15) \quad s_{t+1}g_{t+1} = s_t g_t + \frac{1}{2}\ln\left(1 + \frac{x_t}{1-\rho}\right) + \frac{1}{2}\ln\left(1 - \frac{x_t}{1+\rho}\right).$$

Let us simplify (9.15) further before proceeding. We subtract each side from $s_{t+1}\rho$, using (9.14) to express $s_{t+1}$. This leads to

$$(9.16) \quad \begin{aligned} s_{t+1}x_{t+1} &= s_{t+1}\rho - s_{t+1}g_{t+1} \\ &= s_t x_t - \frac{1}{2}(1-\rho)\ln\left(1 + \frac{x_t}{1-\rho}\right) \\ &\quad - \frac{1}{2}(1+\rho)\ln\left(1 - \frac{x_t}{1+\rho}\right). \end{aligned}$$

Now we update (9.14). For $y \in [0, 2\rho]$, we define

$$f_\rho(y) := \frac{1}{2}\ln\left(1 + \frac{y}{1-\rho}\right) - \frac{1}{2}\ln\left(1 - \frac{y}{1+\rho}\right) - \frac{y}{1-\rho^2},$$

where the inequality $f_\rho(y) \leq 0$ holds since $f_\rho(0) = 0$ and $f'_\rho(y) \leq 0$ for $0 \leq y \leq 2\rho$. Since we consider the algorithm for only $g_t \geq 0$, we have $x_t = \rho - g_t \leq \rho$, so that

$$(9.17) \quad \begin{aligned} s_{t+1} &= s_t + f_\rho(x_t) + \frac{x_t}{1-\rho^2} \\ &\leq s_t + \frac{x_t}{1-\rho^2} = s_t\left(1 + \frac{x_t}{(1-\rho^2)s_t}\right). \end{aligned}$$

We now update (9.16) similarly. We define, for $y \in [0, 2\rho]$,

$$\tilde{f}_\rho(y) := -\frac{1}{2}(1-\rho)\ln\left(1+\frac{y}{1-\rho}\right) - \frac{1}{2}(1+\rho)\ln\left(1-\frac{y}{1+\rho}\right)$$
$$- \frac{y^2}{2(1-\rho^2)} + \frac{2}{3}\frac{\rho y^3}{(1-\rho^2)^2},$$

where the inequality $\tilde{f}_\rho(y) \geq 0$ holds since $\tilde{f}_\rho(0) = 0$ and since one can show $\tilde{f}'_\rho(y) \geq 0$ for $0 \leq y \leq 2\rho$. It thus follows from $x_t \leq \rho$ that

(9.18)
$$x_{t+1}s_{t+1} = x_t s_t + \tilde{f}_\rho(x_t) + \frac{x_t^2}{2(1-\rho^2)} - \frac{2}{3}\frac{\rho x_t^3}{(1-\rho^2)^2}$$
$$\geq x_t s_t \left[1 + \frac{x_t}{2(1-\rho^2)s_t} - \frac{2}{3}\frac{\rho x_t^2}{(1-\rho^2)^2 s_t}\right].$$

Suppose now that

(9.19) $$x_t \leq C t^{-(1/3)-\delta},$$

for $t \geq t_0$, with $\delta > 0$. We can assume, without loss of generality, that $\delta < 2/3$. By (9.17) we then have, for all $t \geq t_0$,

$$s_t = s_{t_0} + \sum_{\ell=t_0}^{t-1}(s_{\ell+1} - s_\ell) \leq s_{t_0} + \sum_{\ell=t_0}^{t-1}\frac{x_\ell}{1-\rho^2} \leq s_{t_0} + \frac{C}{1-\rho^2}\sum_{\ell=t_0}^{t-1}\ell^{-(1/3)-\delta}$$

$$\leq s_{t_0} + \frac{C}{1-\rho^2}\int_{t_0-1}^{t-1} u^{-(1/3)-\delta}\,du \leq s_{t_0} + \frac{C}{1-\rho^2}\frac{(t-1)^{(2/3)-\delta}}{2/3-\delta}.$$

It follows that we can define a finite $C'$ so that for all $t \geq t_0$,

(9.20) $$s_t \leq C' t^{(2/3)-\delta}.$$

Consider now $z_t := x_t^{2-\delta} s_t$. By (9.19) and (9.20) we have, again for $t \geq t_0$,

$$z_t \leq C^{2-\delta}C' t^{(2-\delta)(-(1/3)-\delta)+(2/3)-\delta} = C'' t^{(\delta/3)-2\delta+\delta^2-\delta}$$
$$\leq C'' t^{(\delta/3)-2\delta+2(\delta/3)-\delta} = C'' t^{-2\delta},$$

where we have used that $\delta^2 \leq 2(\delta/3)$ since $\delta < 2/3$. It follows that

(9.21) $$\lim_{t\to\infty} z_t = 0.$$

On the other hand, by (9.17) and (9.18), we have

$$z_{t+1} = x_{t+1}^{2-\delta}s_{t+1} = (x_{t+1}s_{t+1})^{2-\delta}s_{t+1}^{-1+\delta}$$
$$\geq (x_t s_t)^{2-\delta}\left[1 + \frac{x_t}{2(1-\rho^2)s_t} - \frac{2}{3}\frac{\rho x_t^2}{(1-\rho^2)^2 s_t}\right]^{2-\delta}$$
$$\times s_t^{-1+\delta}\left(1 + \frac{x_t}{(1-\rho^2)s_t}\right)^{-1+\delta}.$$

For sufficiently large $t$, $x_t$ will be small so that $x_t(2\rho/3(1-\rho^2)) \leq \delta/4$. Thus,

$$z_{t+1} \geq (x_t s_t)^{2-\delta}\left[1 + \frac{x_t}{2(1-\rho^2)s_t}\left(1 - \frac{\delta}{2}\right)\right]^{2-\delta}$$

(9.22)
$$\times s_t^{-1+\delta}\left(1 + \frac{x_t}{(1-\rho^2)s_t}\right)^{-1+\delta}$$

$$= z_t\left[1 + \frac{x_t}{2(1-\rho^2)s_t}\left(1 - \frac{\delta}{2}\right)\right]^{2-\delta}\left(1 + \frac{x_t}{(1-\rho^2)s_t}\right)^{-1+\delta}.$$

Now consider the function $\phi_\delta(y) = [1 + \frac{y}{2}(1-\frac{\delta}{2})]^{2-\delta}(1+y)^{-1+\delta}$. Since $\phi_\delta(0) = 1$ and $\phi'_\delta(y) = 4^{-2+\delta}[4+y(2-\delta)]^{1-\delta}(1+y)^{-2+\delta}[2y - y\delta + \delta^2]$, it follows that, for sufficiently small $y$,

$$\phi_\delta(y) \geq 1 + \frac{1}{2}\phi'_\delta(0)y = 1 + \frac{\delta^2}{8}y.$$

Since $x_t \to 0$, we have $\lim_{t\to\infty} x_t/s_t = 0$. It then follows from (9.22) that

$$z_{t+1} \geq z_t\left(1 + \frac{\delta^2}{8}\left(\frac{x_t}{(1-\rho^2)s_t}\right)\right)$$

for sufficiently large $t$. This implies $z_{t+1} > z_t$ if $x_t > 0$, but we always have $x_t > 0$ by (3.2). Consequently, there exists a threshold $t_1$ so that $z_t$ is strictly increasing for $t \geq t_1$. Together with $z_{t_1} = s_{t_1} x_{t_1}^{2-\delta} > 0$ (again because $x_{t_1}$ must be nonzero), this contradicts (9.21). It follows that the assumption (9.19) must be false, which completes the proof. $\square$

PROOF OF THEOREM 6.1. We drop the superscripts, since all variables $(\lambda_t, g_t, s_t, \mu_t)$ will be for arc-gv. In order to prove the convergence rate, we need to show that versions of Lemmas 5.1 and 5.3 hold for arc-gv, starting with Lemma 5.1. We have, since tanh can be lower bounded as before, and since for arc-gv we have $\tanh(\gamma_t - \alpha_t^{\text{arc}}) = \mu_t$,

$$\int_{\gamma_t - \alpha_t^{\text{arc}}}^{\gamma_t} \tanh u\, du \geq \tfrac{1}{2}\alpha_t^{\text{arc}}[\tanh \gamma_t + \tanh(\gamma_t - \alpha_t^{\text{arc}})]$$

$$= \tfrac{1}{2}\alpha_t^{\text{arc}}(r_t + \mu_t) \geq \tfrac{1}{2}\alpha_t^{\text{arc}}(r_t + g_t).$$

Using the recursive equation (4.2) with arc-gv's update and simplifying as in the proof of Lemma 5.1 yields the analogous result

$$g_{t+1} - g_t \geq \frac{\alpha_t^{\text{arc}}(r_t - g_t)}{2s_{t+1}}.$$

Since the right-hand side is nonnegative, the sequence of $g_t$'s is nonnegative and nondecreasing; arc-gv makes progress according to the smooth margin.

The proof of Lemma 5.3 follows from only the recursive equation (4.2) and the nonnegativity of the $g_t$'s, so it also holds for arc-gv.

Now we adapt the proof of Theorem 5.2. Since we have just shown that the statements of Lemmas 5.1 and 5.3 both hold for arc-gv, we can exactly use the proof of Theorem 5.2 from the beginning through equation (9.9); we must then specialize to arc-gv. We define $\Delta\mu(\boldsymbol{\lambda}_t) = \rho - \mu_t$,

$$\alpha_t^{\mathrm{arc}} \geq \tanh \alpha_t^{\mathrm{arc}} = \tanh[\gamma_t - (\gamma_t - \alpha_t^{\mathrm{arc}})] = \frac{\tanh \gamma_t - \tanh(\gamma_t - \alpha_t^{\mathrm{arc}})}{1 - \tanh \gamma_t \tanh(\gamma_t - \alpha_t^{\mathrm{arc}})}$$

$$= \frac{r_t - \mu_t}{1 - r_t \mu_t} \geq \frac{\rho - \mu_t}{1} = \Delta\mu(\boldsymbol{\lambda}_t).$$

Thus, we have

$$s_{t+1} = s_{\tilde{1}} + \sum_{\ell=\tilde{1}}^{t} \alpha_\ell \geq s_{\tilde{1}} + \sum_{\ell=\tilde{1}}^{t} \Delta\mu(\boldsymbol{\lambda}_\ell) \geq s_{\tilde{1}} + (t - \tilde{1} + 1) \min_{\ell \in 1,\ldots,t} \Delta\mu(\boldsymbol{\lambda}_\ell),$$

or, changing the index and using $\min_{\ell \in 1,\ldots,t-1} \Delta\mu(\boldsymbol{\lambda}_\ell) \geq \min_{\ell \in 1,\ldots,t} \Delta\mu(\boldsymbol{\lambda}_\ell)$,

$$s_t \geq s_{\tilde{1}} + (t - \tilde{1}) \min_{\ell \in 1,\ldots,t} \Delta\mu(\boldsymbol{\lambda}_\ell).$$

Combining with (9.9), using $\Delta G(\boldsymbol{\lambda}_t) \geq \Delta\mu(\boldsymbol{\lambda}_t) \geq \min_{\ell \in 1,\ldots,t} \Delta\mu(\boldsymbol{\lambda}_\ell)$,

$$t - \tilde{1} \leq \frac{s_t}{\min_{\ell \in 1,\ldots,t} \Delta\mu(\boldsymbol{\lambda}_\ell)} \leq \frac{(s_{\tilde{1}} + \ln 2)[\Delta G(\boldsymbol{\lambda}_{\tilde{1}})]^{2/(1-\rho)}}{[\min_{\ell \in 1,\ldots,t} \Delta\mu(\boldsymbol{\lambda}_\ell)]^{[1+2/(1-\rho)]}},$$

which means that $\min_{\ell \in 1,\ldots,t} \Delta\mu(\boldsymbol{\lambda}_\ell) \geq \varepsilon$ is possible only if

$$t \leq \tilde{1} + (s_{\tilde{1}} + \ln 2)\varepsilon^{-(3-\rho)/(1-\rho)}.$$

If $t$ exceeds this value, $\min_{\ell \in 1,\ldots,t} \Delta\mu(\boldsymbol{\lambda}_\ell) < \varepsilon$. This concludes the proof. $\square$

## 10. Proofs from Section 7.

PROOF OF THEOREM 7.2. We drop the superscripts [A] during this proof. We need to show that $g_{t+1} \geq g_t$ for all $t$ implies that $r_t \to r$ and that $g_t \to \Upsilon(r)$. Using the argument of Theorem 7.1, an increase in $G$ means that $\Upsilon(r_t) = g_t + c_t$ where $c_t > 0$. Equivalently, by (4.2) and the definition of $\Upsilon(r_t)$,

$$s_{t+1}g_{t+1} = \Upsilon(r_t)\alpha_t + s_t g_t = (g_t + c_t)\alpha_t + s_t g_t = s_{t+1}g_t + c_t\alpha_t,$$

and dividing by $s_{t+1}$ we have

$$g_{t+1} = g_t + \frac{c_t \alpha_t}{s_{t+1}}.$$

We need to show that $c_t \to 0$.

We are interested only in the later iterations, where AdaBoost is "close" to the cycle. To ease notation, without loss of generality we will assume that at $t=1$, AdaBoost is already close to the cycle. More precisely, we assume that for some $\varepsilon_\alpha > 0$, for all integers $a \geq 0$, for all $0 \leq k < T$ (excluding $a=0$, $k=0$ since $t$ starts at 1),

$$\alpha_{aT+k} \geq \alpha_{\text{lowerbd},k}, \qquad \text{where } \alpha_{\text{lowerbd},k} := \left(\lim_{\bar{a}\to\infty} \alpha_{\bar{a}T+k}\right) - \varepsilon_\alpha > 0.$$

Also, for some $\varepsilon_s > 0$, for all integers $a \geq 1$, for all $0 \leq k < T$, we assume

$$s_{aT+k} \leq as_{\text{upperbd}} + s_k, \qquad \text{where } s_{\text{upperbd}} \geq \sum_{\bar{k}=0}^{T-1}\left(\lim_{\bar{a}\to\infty} \alpha_{\bar{a}T+\bar{k}}\right) + \varepsilon_s.$$

Since AdaBoost is converging to a cycle, we know that $r_t$ is not much different from its limiting value, that is, that for any arbitrarily small positive $\varepsilon_\Upsilon$ there exists $T_{\varepsilon_\Upsilon}$ such that $t > T_{\varepsilon_\Upsilon}$ implies

$$\left|\Upsilon(r_t) - \lim_{a\to\infty}\Upsilon(r_{t+aT})\right| < \varepsilon_\Upsilon.$$

This implies $\Upsilon(r_{t-T}) > \Upsilon(r_t) - 2\varepsilon_\Upsilon$ for $t > T_{\varepsilon_\Upsilon} + T$. This also implies $\Upsilon(r_{t-2T}) > \Upsilon(r_t) - 2\varepsilon_\Upsilon$ for $t > T_{\varepsilon_\Upsilon} + 2T$, and so on. Let us first choose an arbitrarily small value for $\varepsilon_\Upsilon$. Accordingly, find an iteration $\tilde{t} > T_{\varepsilon_\Upsilon} + T$ so that $c_{\tilde{t}} > 2\varepsilon_\Upsilon > 0$. (If $\tilde{t}$ does not exist for any $\varepsilon_\Upsilon$, the result is trivial since we automatically have $c_t \to 0$, which we are trying to prove.)

First we will show that there is a strict increase in $G$ at the same point in previous cycles. Since $G$ is nondecreasing by our assumption, we have $g_{\tilde{t}} \geq g_{\tilde{t}-T}$. Thus $\Upsilon(r_{\tilde{t}}) = g_{\tilde{t}} + c_{\tilde{t}} \geq g_{\tilde{t}-T} + c_{\tilde{t}}$. Hence,

$$\Upsilon(r_{\tilde{t}-T}) \geq \Upsilon(r_{\tilde{t}}) - 2\varepsilon_\Upsilon = g_{\tilde{t}} + c_{\tilde{t}} - 2\varepsilon_\Upsilon \geq g_{\tilde{t}-T} + c_{\tilde{t}} - 2\varepsilon_\Upsilon.$$

Thus, a strict increase occurred at time $\tilde{t}-T$ as well, with $c_{\tilde{t}-T} \geq c_{\tilde{t}} - 2\varepsilon_\Upsilon > 0$. Let us repeat exactly this argument for $\tilde{t}-2T$: since $G$ is nondecreasing, $g_{\tilde{t}} \geq g_{\tilde{t}-2T}$. Thus a strict increase in $G$ at $\tilde{t}$ implies

$$\Upsilon(r_{\tilde{t}-2T}) \geq \Upsilon(r_{\tilde{t}}) - 2\varepsilon_\Upsilon = g_{\tilde{t}} + c_{\tilde{t}} - 2\varepsilon_\Upsilon \geq g_{\tilde{t}-2T} + c_{\tilde{t}} - 2\varepsilon_\Upsilon.$$

So a strict increase occurred at time $\tilde{t}-2T$ with $c_{\tilde{t}-2T} \geq c_{\tilde{t}} - 2\varepsilon_\Upsilon > 0$. Continuing to repeat this argument for past cycles shows that if $c_{\tilde{t}} > 2\varepsilon_\Upsilon > 0$, then $c_{\tilde{t}-T} > 0$, $c_{\tilde{t}-2T} > 0$, $c_{\tilde{t}-3T} > 0$, for iterations at least as far back as $T_{\varepsilon_\Upsilon}$. What we have shown is that a strict increase in $G$ implies a strict increase in $G$ at the same point in previous cycles. Let us show the theorem by contradiction. We make the weakest possible assumption: for some large $t$, a strict increase in $G$ occurs (hence a strict increase occurs at the same point in a previous cycle). These iterations where the increase occurs are assumed without loss of generality to be $aT$, where $a \in \{1,2,3,\ldots\}$. (If $T_{\varepsilon_\Upsilon} > 1$, we simply renumber the iterations to ease notation.) For all other iterations, $G$ is assumed only to be nondecreasing. We need to show $\lim_{\bar{a}\to\infty} c_{\bar{a}T} = 0$. We now have for $a > 1$,

$$g_{aT} \geq g_{(a-1)T+1} = g_{(a-1)T} + \frac{c_{(a-1)T}\alpha_{(a-1)T}}{s_{(a-1)T+1}} \geq g_T + \sum_{\bar{a}=1}^{a-1} \frac{c_{\bar{a}T}\alpha_{\bar{a}T}}{s_{\bar{a}T+1}}.$$

Putting this together with $s_{aT+k} \leq a s_{\text{upperbd}} + s_k$ and $\alpha_{aT+k} \geq \alpha_{\text{lowerbd},k}$, we find that

$$g_{aT} \geq g_T + \sum_{\bar{a}=1}^{a-1} \frac{c_{\bar{a}T}\alpha_{\text{lowerbd},0}}{\bar{a} s_{\text{upperbd}} + s_1}.$$

Since $s_{\text{upperbd}}$ and $\alpha_{\text{lowerbd},0}$ are constants, the partial sums become arbitrarily large if no infinite subsequence of the $c_{\bar{a}T}$'s approaches zero. So, there exists a subsequence $1', 2', 3', \ldots$ such that $\lim_{a'} c_{a'T} = 0$. Considering only this subsequence, and taking the limits of both sides of the equation $\Upsilon(r_{a'T}) = g_{a'T} + c_{a'T}$, we obtain

(10.1) $$\lim_{a' \to \infty} \Upsilon(r_{a'T}) = \lim_{a' \to \infty} g_{a'T}.$$

Since AdaBoost is assumed to be converging to a cycle and since $1'T, 2'T, 3'T, \ldots$ is a subsequence of $T, 2T, 3T, \ldots$, then $r := \lim_{a' \to \infty} r_{a'T}$ exists. Thus,

(10.2) $$\lim_{a' \to \infty} \Upsilon(r_{a'T}) = \Upsilon(r) = \lim_{a \to \infty} \Upsilon(r_{aT}).$$

Now, since $G$ is a monotonically increasing sequence that is bounded by 1,

(10.3) $$\lim_{t' \to \infty} g_{a'T} = \lim_{t \to \infty} g_t = \lim_{a \to \infty} g_{aT}.$$

Recall that by definition, $\Upsilon(r_{aT}) - g_{aT} = c_{aT}$. Taking the limit of both sides as $a \to \infty$, and using (10.1), (10.2) and (10.3), we find

$$0 = \lim_{a \to \infty} [\Upsilon(r_{aT}) - g_{aT}] = \lim_{a \to \infty} c_{aT}.$$

Thus, even if we make the weakest possible assumption, namely that there is a strict increase even once per cycle, the increase goes to zero. In other words, our initial assumption was that the $c_{aT}$'s are strictly positive (not prohibiting other $c_t$'s from being positive as well), and we have shown that their limit must be zero. So we cannot have strict increases at all, $c_t \to 0$. Thus, we must have

$$0 = \lim_{t \to \infty} c_t = \lim_{t \to \infty} [\Upsilon(r_t) - g_t], \qquad \text{so } \lim_{t \to \infty} g_t = \lim_{t \to \infty} \Upsilon(r_t) = \Upsilon(r).$$

This means all $r_t$'s in the cycle are identical, $r_t \to r$. We have finished the proof. $\square$

PROOF OF THEOREM 7.4. Again we drop superscripts $^{[A]}$. Choose $\delta > 0$ arbitrarily small. We shall prove that $\limsup_t g_t \leq \Upsilon(\bar{\rho}+\sigma)+\delta$ and $\liminf_t g_t \geq \Upsilon(\bar{\rho})-\delta$, which (since $\delta$ was arbitrarily small) would prove the theorem. We start with the recursive equation (4.2). Subtracting $\alpha_t g_t$ from both sides and simplifying yields $s_{t+1}(g_{t+1} - g_t) = \Upsilon(r_t)\alpha_t - \alpha_t g_t$, and dividing by $s_{t+1}$,

$$(10.4) \qquad g_{t+1} - g_t = (\Upsilon(r_t) - g_t)\frac{\alpha_t}{s_{t+1}}.$$

First we will show that, for some $t$, if $g_t$ is smaller than $\Upsilon(\bar{\rho}) - \delta$, then $g_t$ must monotonically increase for $\tilde{t} \geq t$ until $g_{\tilde{t}}$ meets $\Upsilon(\bar{\rho}) - \delta$ after a finite number of steps. Suppose $g_t$ is smaller than $\Upsilon(\bar{\rho}) - \delta$, and moreover suppose this is true for $N$ iterations: $\Upsilon(\bar{\rho}) - g_{\tilde{t}} > \delta > 0$, for $\tilde{t} \in \{t, t+1, t+2, \ldots, t+N\}$. Then, since $\Upsilon(r_{\tilde{t}}) \geq \Upsilon(\bar{\rho})$, we have

$$g_{\tilde{t}+1} - g_{\tilde{t}} > \delta \frac{\alpha_{\tilde{t}}}{s_{\tilde{t}+1}} \geq \delta \frac{\tanh^{-1}\bar{\rho}}{\tanh^{-1}(\bar{\rho}+\sigma)}\frac{1}{\tilde{t}+1} > 0,$$

where we have used that $\alpha_{\tilde{t}} = \tanh^{-1} r_{\tilde{t}} \geq \tanh^{-1}\bar{\rho}$ and $s_{\tilde{t}+1} \leq (\tilde{t}+1)\tanh^{-1}(\bar{\rho}+\sigma)$, which are due to the restrictions on $r_t$. Recursion yields

$$g_{t+N} - g_t \geq \delta \frac{\tanh^{-1}\bar{\rho}}{\tanh^{-1}(\bar{\rho}+\sigma)}\left[\frac{1}{t+1} + \frac{1}{t+2} + \cdots + \frac{1}{t+N}\right],$$

$$\geq \delta \frac{\tanh^{-1}\bar{\rho}}{\tanh^{-1}(\bar{\rho}+\sigma)}\int_{t+1}^{t+N+1}\frac{1}{x}dx$$

$$= \delta \frac{\tanh^{-1}\bar{\rho}}{\tanh^{-1}(\bar{\rho}+\sigma)}\ln\left(1 + \frac{N}{t+1}\right).$$

Because $1 \geq g_{t+N} - g_t$, this implies

$$N \leq (t+1)\exp\left[\frac{1}{\delta}\frac{\tanh^{-1}(\bar{\rho}+\sigma)}{\tanh^{-1}\bar{\rho}}\right] =: N_t.$$

It follows that there must be at least one value $N$ in $\{0, 1, 2, \ldots, N_t, N_t+1\}$ such that $\Upsilon(\bar{\rho}) - g_{t+N} \leq \delta$.

An identical argument can be made to show that if $g_t - \Upsilon(\bar{\rho}+\sigma) > \delta > 0$, then the values of $g_{\tilde{t}}$, for $\tilde{t} \geq t$ will monotonically decrease to meet $\Upsilon(\bar{\rho}+\sigma) + \delta$. To make this explicit, suppose that $g_{\tilde{t}} - \Upsilon(\bar{\rho}+\sigma) > \delta > 0$ for $\tilde{t} \in \{t, t+1, \ldots, t+M\}$. Then, since $-\Upsilon(r_{\tilde{t}}) \geq -\Upsilon(\bar{\rho}+\sigma)$,

$$g_{\tilde{t}} - g_{\tilde{t}+1} = (g_{\tilde{t}} - \Upsilon(r_{\tilde{t}}))\frac{\alpha_{\tilde{t}}}{s_{\tilde{t}+1}} \geq \delta \frac{\tanh^{-1}\bar{\rho}}{\tanh^{-1}(\bar{\rho}+\sigma)}\frac{1}{\tilde{t}+1}.$$

By the same reasoning as above, it follows that $M$ cannot exceed some finite $M_t$. Therefore, we must have, for some $\tilde{t} \in \{t+1, \ldots, t+M_t, t+M_t+1\}$, that $g_{\tilde{t}} - \Upsilon(\bar{\rho} + \sigma) \leq \delta$, and that $g_t$ decreases monotonically until this condition is met.

To summarize, we have just shown that the sequence of values of $g_t$ cannot remain below $\Upsilon(\bar{\rho}) - \delta$, and cannot remain above $\Upsilon(\bar{\rho} + \sigma) + \delta$. Next we show that from some $t_0$ onward, the $g_t$'s cannot even leave the interval $[\Upsilon(\bar{\rho}) - \delta, \Upsilon(\bar{\rho} + \sigma) + \delta]$. First of all, note that we can upper bound $|g_{t+1} - g_t|$, regardless of its sign, as follows:

$$|g_{t+1} - g_t| = |\Upsilon(r_t) - g_t| \frac{\alpha_t}{s_{t+1}}$$

$$\leq \max(\Upsilon(\bar{\rho} + \sigma), 1) \frac{\tanh^{-1}(\bar{\rho} + \sigma)}{\tanh^{-1}\bar{\rho}} \frac{1}{t+1} =: C_\sigma \frac{1}{t+1},$$

where we have used $|\Upsilon(r_t) - g_t| \leq \max(\Upsilon(r_t), g_t) \leq \max(\Upsilon(\bar{\rho} + \sigma), 1)$, since $\Upsilon(r_t)$ and $g_t$ are both positive and bounded.

Now, if $t \geq C_\sigma[\Upsilon(\bar{\rho} + \sigma) - \Upsilon(\bar{\rho}) + \delta]^{-1} =: T_1$, then the bound we just proved implies that the $g_t$ for $t \geq T_1$ cannot jump from values below $\Upsilon(\bar{\rho}) - \delta$ to values above $\Upsilon(\bar{\rho} + \sigma) + \delta$ in one time step. Since we know that the $g_t$ cannot remain below $\Upsilon(\bar{\rho}) - \delta$ or above $\Upsilon(\bar{\rho}) + \delta$ for more than $\max(N_t, M_t)$ consecutive steps, it follows that for $t \geq T_1$, the $g_t$ must return to $[\Upsilon(\bar{\rho}) - \delta, \Upsilon(\bar{\rho} + \sigma) + \delta]$ infinitely often. Pick $t_0 \geq T_1$ so that $g_{t_0} \in [\Upsilon(\bar{\rho}) - \delta, \Upsilon(\bar{\rho} + \sigma) + \delta]$. We distinguish three cases: $g_{t_0} < \Upsilon(\bar{\rho})$, $\Upsilon(\bar{\rho}) \leq g_{t_0} \leq \Upsilon(\bar{\rho} + \sigma)$ and $g_{t_0} > \Upsilon(\bar{\rho} + \sigma)$. In the first case, we know from (10.4) that $g_{t_0+1} - g_{t_0} > 0$, so that

$$g_{t_0} < g_{t_0+1} \leq g_{t_0} + C_\sigma \frac{1}{t_0 + 1} \leq \Upsilon(\bar{\rho}) + \Upsilon(\bar{\rho} + \sigma) - \Upsilon(\bar{\rho}) + \delta,$$

that is, $g_{t_0+1} \in [\Upsilon(\bar{\rho}) - \delta, \Upsilon(\bar{\rho} + \sigma) + \delta]$. A similar argument applies to the third case. In the middle case, we find that

$$\text{dist}(g_{t_0+1}, [\Upsilon(\bar{\rho}), \Upsilon(\bar{\rho} + \sigma)]) := \max(0, g_{t_0+1} - \Upsilon(\bar{\rho} + \sigma), \Upsilon(\bar{\rho}) - g_{t_0+1})$$

$$\leq |g_{t_0+1} - g_{t_0}| \leq \frac{C_\sigma}{t_0 + 1},$$

which does not exceed $\delta$ if $t_0 \geq C_\sigma \delta^{-1} =: T_2$. It follows that if $t_0 \geq T_0 := \max(T_1, T_2)$, and $g_{t_0} \in [\Upsilon(\bar{\rho}) - \delta, \Upsilon(\bar{\rho} + \sigma) + \delta]$, then $g_{t_0+1}$ will likewise be in $[\Upsilon(\bar{\rho}) - \delta, \Upsilon(\bar{\rho} + \sigma) + \delta]$. By induction we obtain that $g_t \in [\Upsilon(\bar{\rho}) - \delta, \Upsilon(\bar{\rho} + \sigma) + \delta]$ for all $t \geq t_0$. This implies

$$\liminf_{t \to \infty} g_t \geq \Upsilon(\bar{\rho}) - \delta \quad \text{and} \quad \limsup_{t \to \infty} g_t \leq \Upsilon(\bar{\rho} + \sigma) + \delta.$$

Since, at the start of this proof, $\delta > 0$ could be chosen arbitrarily small, we obtain $\liminf_{t \to \infty} g_t \geq \Upsilon(\bar{\rho})$ and $\limsup_{t \to \infty} g_t \leq \Upsilon(\bar{\rho} + \sigma)$.

Note that we do not really need uniform bounds on $r_t$ for this proof to work. In fact, we need only bounds that hold "eventually," so it is sufficient that $\limsup_t r_t \leq \bar{\rho} + \sigma$, $\liminf_t r_t \geq \bar{\rho}$. In the special case where $\lim_t r_t = \rho$, that is, where $\sigma = 0$ and $\bar{\rho} = \rho$, it then follows that $\lim_t g_t = \Upsilon(\rho)$. Hence we have completed the proof. □

PROOF OF THEOREM 7.5. For any given $\bar{\rho}$ and $\sigma$, we will create a matrix $\mathbf{M}$ such that edge values can always be chosen within $[\bar{\rho}, \bar{\rho} + \sigma]$. For this matrix $\mathbf{M}$, we must also have $\bar{\rho} \geq \rho$. Choose a value for $\bar{\rho}$, and choose $\sigma$ arbitrarily small. Also, for reasons that will become clear later, choose a constant $\phi$ such that

$$\phi \geq \frac{1 + \bar{\rho} + \sigma}{1 - \bar{\rho} - \sigma},$$

and choose $m \geq 2\phi/\sigma$. As usual, $m$ will be the number of training examples. Let $\mathbf{M}$ contain only the set of possible columns that have at most $m(\bar{\rho}+1)/2$ entries that are $+1$. (We can assume $m$ was chosen so that this is an integer.) This completes our construction of $\mathbf{M}$.

Before we continue, we need to prove that for $\rho$ of this matrix $\mathbf{M}$, we have $\rho \leq \bar{\rho}$. For any column $j$,

$$\sum_{i=1}^{m} M_{ij} \leq (+1)\frac{m(\bar{\rho}+1)}{2} + (-1)\left(m - \frac{m(\bar{\rho}+1)}{2}\right) = m\bar{\rho}.$$

Thus, for any $\bar{\boldsymbol{\lambda}} \in \Delta_n$, we upper bound the average margin (i.e., the average margin over training examples),

$$\frac{1}{m}\sum_{i=1}^{m}\sum_{j=1}^{n} \bar{\lambda}_j M_{ij} = \sum_j \bar{\lambda}_j \left(\frac{1}{m}\sum_{i=1}^{m} M_{ij}\right) \leq \sum_j \bar{\lambda}_j \frac{1}{m} m\bar{\rho} = \bar{\rho}\sum_j \bar{\lambda}_j = \bar{\rho}.$$

We have just shown that the average margin is at most $\bar{\rho}$. There must be at least one training example that achieves a margin at or below the average margin; thus $\min_i(\mathbf{M}\bar{\boldsymbol{\lambda}})_i \leq \bar{\rho}$, and since $\bar{\boldsymbol{\lambda}}$ is arbitrary, $\rho = \max_{\bar{\boldsymbol{\lambda}} \in \Delta_n} \min_i(\mathbf{M}\bar{\boldsymbol{\lambda}})_i \leq \bar{\rho}$, the maximum margin is at most $\bar{\rho}$.

We will now describe our procedure for choosing weak classifiers, and then prove that this procedure always chooses edge values $r_t$ within $[\bar{\rho}, \bar{\rho} + \sigma]$. As usual, for $t = 1$ we set $d_{1,i} = 1/m$ for all $i$. Let us describe the procedure to choose our weak classifier $j_t$, for iteration $t$. Without loss of generality, we reorder the training examples so that $d_{t,1} \geq d_{t,2} \geq \cdots \geq d_{t,m}$, for convenience of notation in describing the procedure. We choose a weak classifier $j_t$ that correctly classifies the first $\bar{i}$ training examples, where $\bar{i}$ is the smallest index such that $2(\sum_{i=1}^{\bar{i}} d_{t,i}) - 1 \geq \bar{\rho}$. That is, we correctly classify enough examples so that the edge just exceeds $\bar{\rho}$. The maximum number of correctly classified examples, $\bar{i}$, will be at most $m(\bar{\rho}+1)/2$, corresponding to the case where $d_{t,1} = \cdots = d_{t,m} = 1/m$. Thus, the weak classifier we choose thankfully corresponds to a column of $\mathbf{M}$. The edge $r_t$ is $r_t = 2(\sum_{i=1}^{\bar{i}} d_{t,i}) - 1 \geq \bar{\rho}$. We can now update AdaBoost's weight vector using the usual exponential rule. Thus, our description of the procedure is complete.

By definition, we have chosen the edge such that $\bar{\rho} \leq r_t$. We have only to show that $r_t \leq \bar{\rho} + \sigma$ for each $t$. The main step in our proof is to show that $\phi = K_1 = K_t$ for all $t$, where for each iteration $t$,

$$K_t := \max\left\{\max_{i_1, i_2} \frac{d_{t,i_1}}{d_{t,i_2}}, \phi\right\}.$$

We will prove this by induction. For the base case $t = 1$, $K_1 = \max\{1, \phi\} = \phi$. Now for the inductive step. In order to make calculations easier, we will write AdaBoost's weight update in a different way (this iterated map can be derived from the usual exponential update) [22, 23]. Namely,

$$d_{t+1,i} = \begin{cases} \dfrac{d_{t,i}}{1+r_t}, & \text{for } i \leq \bar{i}, \\ \dfrac{d_{t,i}}{1-r_t}, & \text{for } i > \bar{i}. \end{cases}$$

Assuming $\phi = K_t$, we will show that $K_{t+1} = K_t$. We can calculate the value of $K_{t+1}$ using the update rule written above,

$$K_{t+1} = \max\left\{\max_{i_1, i_2} \frac{d_{t+1,i_1}}{d_{t+1,i_2}}, \phi\right\} = \left\{\frac{\max_{i_1} d_{t+1,i_1}}{\min_{i_2} d_{t+1,i_2}}, \phi\right\}$$

$$= \max\left\{\frac{\max\{d_{t,1}/(1+r_t), d_{t,\bar{i}+1}/(1-r_t)\}}{\min\{d_{t,\bar{i}}/(1+r_t), d_{t,m}/(1-r_t)\}}, \phi\right\}$$

$$= \max\left\{\frac{d_{t,1}}{d_{t,\bar{i}}}, \frac{d_{t,\bar{i}+1}}{d_{t,m}}, \frac{d_{t,1}}{d_{t,m}} \frac{1-r_t}{1+r_t}, \frac{d_{t,\bar{i}+1}}{d_{t,\bar{i}}} \frac{1+r_t}{1-r_t}, \phi\right\}.$$

By our inductive assumption, the ratios of $d_{t,i}$ values are all nicely bounded, that is, $\frac{d_{t,1}}{d_{t,\bar{i}}} \leq K_t = \phi$, $\frac{d_{t,\bar{i}+1}}{d_{t,m}} \leq \phi$ and $\frac{d_{t,1}}{d_{t,m}} \leq \phi$. Another bound we have automatically is $(1-r_t)/(1+r_t) \leq 1$. We have now shown that none of the first three terms can be greater than $\phi$, thus they can be ignored. Consider just the fourth term. Since we have ordered the training examples, $\frac{d_{t,\bar{i}+1}}{d_{t,\bar{i}}} \leq 1$. If we can bound $(1+r_t)/(1-r_t)$ by $\phi$, we will be done with the induction. We can bound the edge $r_t$ from above, using our choice of $\bar{i}$. Namely, we chose $\bar{i}$ so that the edge exceeds $\bar{\rho}$ by the influence of at most one extra training example,

(10.5) $$r_t \leq \bar{\rho} + 2\max_i d_{t,i} = \bar{\rho} + 2d_{t,1}.$$

Let us now upper bound $d_{t,1}$. By definition of $K_t$, we have $\frac{d_{t,1}}{d_{t,m}} \leq K_t$, and thus $d_{t,1} \leq K_t d_{t,m} \leq K_t/m$. Here, we have used that $d_{t,m} = \min_i d_{t,i} \leq 1/m$ since the $\mathbf{d}_t$ vectors are normalized to 1. By our specification that $m \geq 2\phi/\sigma$ and by our induction principle, we have $d_{t,1} \leq K_t/m \leq \phi\sigma/2\phi = \sigma/2$. Using (10.5), $r_t \leq \bar{\rho} + 2\sigma/2 = \bar{\rho} + \sigma$. (This is by design.) So,

$$\frac{1+r_t}{1-r_t} \leq \frac{1+\bar{\rho}+\sigma}{1-\bar{\rho}-\sigma} \leq \phi.$$

Thus, $K_{t+1} = \phi$. We have just shown that for this procedure, $K_t = \phi$ for all $t$.

Lastly, we note that since $K_t = \phi$ for all $t$, we will always have $r_t \leq \bar{\rho} + \sigma$, by the upper bound for $r_t$ we have just calculated. $\square$

**11. Conclusions.** Our broad goal is to understand the generalization properties of boosting algorithms such as AdaBoost. This is a large and difficult problem that has been studied for a decade. Yet, how are we to understand generalization when even the most basic convergence properties of the most commonly used boosting algorithm are not well understood? AdaBoost's convergence properties are understood in precisely two cases, namely the cyclic case, and the case of bounded edges introduced here.

Our work consists of two main contributions, both of which use the smooth margin function as an important tool. First, from the smooth margin itself, we derive and analyze the algorithms coordinate ascent boosting and approximate coordinate ascent boosting. These algorithms are similar to AdaBoost in that they are adaptive and based on coordinate ascent. However, their convergence can be understood, namely, both algorithms converge to a maximum margin solution with a fast convergence rate. We also give an analogous convergence rate for Breiman's arc-gv algorithm. Our second contribution is an analysis of AdaBoost in terms of the smooth margin. We analyze the case where AdaBoost exhibits cyclic behavior, and we present the case of bounded edges. In the case of bounded edges, we are able to derive a direct relationship between AdaBoost's edge values (which measure the performance of the weak learning algorithm) and the asymptotic margin.

11.1. *Open problems.* We leave open a long list of relevant problems. We have made much progress in understanding AdaBoost's convergence in general via the understanding of special cases, such as the cyclic setting and the setting with bounded edges. The next interesting questions are even more general; for a given matrix **M**, can we predict whether optimal-case AdaBoost will converge to a maximum margin solution? Also, is there a procedure for choosing weak classifiers in the nonoptimal case that would always force convergence to a maximum margin solution? In this case, one would have to plan ahead in order to attain large edge values.

Another open area involves numerical experiments; our new algorithms fall "in between" AdaBoost and arc-gv in many ways; for example, our new algorithms have step sizes that are in between arc-gv and AdaBoost. Can we determine which problem domains match with which algorithms? From our experiments, we suspect the answer to this is quite subtle, and in many domains, all of these algorithms may be tied (within some error precision).

We have presented a controlled numerical experiment using only AdaBoost, to show that the weak learning algorithm (and thus the margin) may have a large impact on generalization. Other experiments along the same lines can be suggested; for example, if the weak learning algorithm is simply bounded from above (cannot choose an edge above $c$ where $0 \ll c < 1$), does this restriction limit the generalization ability of the algorithm? From our convergence analysis, it is clear that this sort of limitation might yield clarity in convergence calculations, considering that a significant portion of our convergence calculations are step-size bounds.

**Acknowledgments.** Thanks to Manfred Warmuth, Gunnar Rätsch and our anonymous reviewers for helpful comments.

C. Rudin  
Center for Computational Learning Systems  
Columbia University  
Interchurch Center  
475 Riverside MC 7717  
New York, New York 10115  
USA  
E-mail: rudin@ccls.columbia.edu

R. E. Schapire  
Department of Computer Science  
Princeton University  
35 Olden St.  
Princeton, New Jersey 08544  
USA  
E-mail: schapire@cs.princeton.edu

I. Daubechies  
Program in Applied and Computational Mathematics  
Princeton University  
Fine Hall, Washington Road  
Princeton, New Jersey 08544-1000  
USA  
E-mail: ingrid@math.princeton.edu